\documentclass[letterpaper, 10 pt, conference]{ieeeconf}  

\IEEEoverridecommandlockouts                              

\makeatletter
\let\NAT@parse\undefined
\makeatother

\usepackage{cite}
\usepackage{amsmath,amssymb,amsfonts}
\usepackage{algorithmic}
\usepackage{graphicx}
\usepackage{pifont}
\usepackage{textcomp}
\usepackage{hyperref}
\usepackage{multirow}
\usepackage{balance}
\usepackage[table,xcdraw]{xcolor}

\def\rot{\rotatebox} 

\newcommand{\cmark}{\ding{51}} 
\newcommand{\xmark}{\ding{55}} 
\newcommand{\smark}{\ding{72}} 

\title{\LARGE \bf
AutoDRIVE: A Comprehensive, Flexible and Integrated Digital Twin Ecosystem for Enhancing Autonomous Driving Research and Education
}

\author{Tanmay Samak$^{\ast 1, 2}$, Chinmay Samak$^{\ast 1, 2}$, Sivanathan Kandhasamy$^{1}$, Venkat Krovi$^{2}$ and Ming Xie$^{3}$
\thanks{$^{\ast}$These authors contributed equally.}
\thanks{$^{1}$Autonomous Systems Lab (ASL), Department of Mechatronics Engineering, SRM Institute of Science and Technology (SRMIST), Kattankulathur, Tamil Nadu 603203, India.
{\tt\small {\{\href{mailto:tv4813@srmist.edu.in}{tv4813}, \href{mailto:cv4703@srmist.edu.in}{cv4703}, \href{mailto:sivanatk@srmist.edu.in}{sivanatk}\}@srmist.edu.in}}}%
\thanks{$^{2}$Automation, Robotics and Mechatronics Lab (ARMLab), Department of Automotive Engineering, Clemson University International Center for Automotive Research (CU-ICAR), Greenville, SC 29607, USA.
{\tt\small {\{\href{mailto:tsamak@clemson.edu}{tsamak}, \href{mailto:csamak@clemson.edu}{csamak}, \href{mailto:vkrovi@clemson.edu}{vkrovi}\}@clemson.edu}}}%
\thanks{$^{3}$School of Mechanical and Aerospace Engineering, Nanyang Technological University, Singapore 639798, Singapore.
{\tt\small \href{mailto:mmxie@ntu.edu.sg}{mmxie@ntu.edu.sg}}}%
}

\begin{document}

\maketitle
\thispagestyle{empty}
\pagestyle{empty}


\begin{abstract}
Prototyping and validating hardware-software components, sub-systems and systems within the intelligent transportation system-of-systems framework requires a modular yet flexible and open-access ecosystem. This work presents our attempt towards developing such a comprehensive research and education ecosystem, called AutoDRIVE, for synergistically prototyping, simulating and deploying cyber-physical solutions pertaining to autonomous driving as well as smart city management. AutoDRIVE features both software as well as hardware-in-the-loop testing interfaces with openly accessible scaled vehicle and infrastructure components. The ecosystem is compatible with a variety of development frameworks, and supports both single and multi-agent paradigms through local as well as distributed computing. Most critically, AutoDRIVE is intended to be modularly expandable to explore emergent technologies, and this work highlights various complementary features and capabilities of the proposed ecosystem by demonstrating four such deployment use-cases: (i) autonomous parking using probabilistic robotics approach for mapping, localization, path planning and control; (ii) behavioral cloning using computer vision and deep imitation learning; (iii) intersection traversal using vehicle-to-vehicle communication and deep reinforcement learning; and (iv) smart city management using vehicle-to-infrastructure communication and internet-of-things.\\%
\end{abstract}


\section{Introduction}
\label{Section: Introduction}

\begin{figure}[t]
\centering
\includegraphics[width=\linewidth]{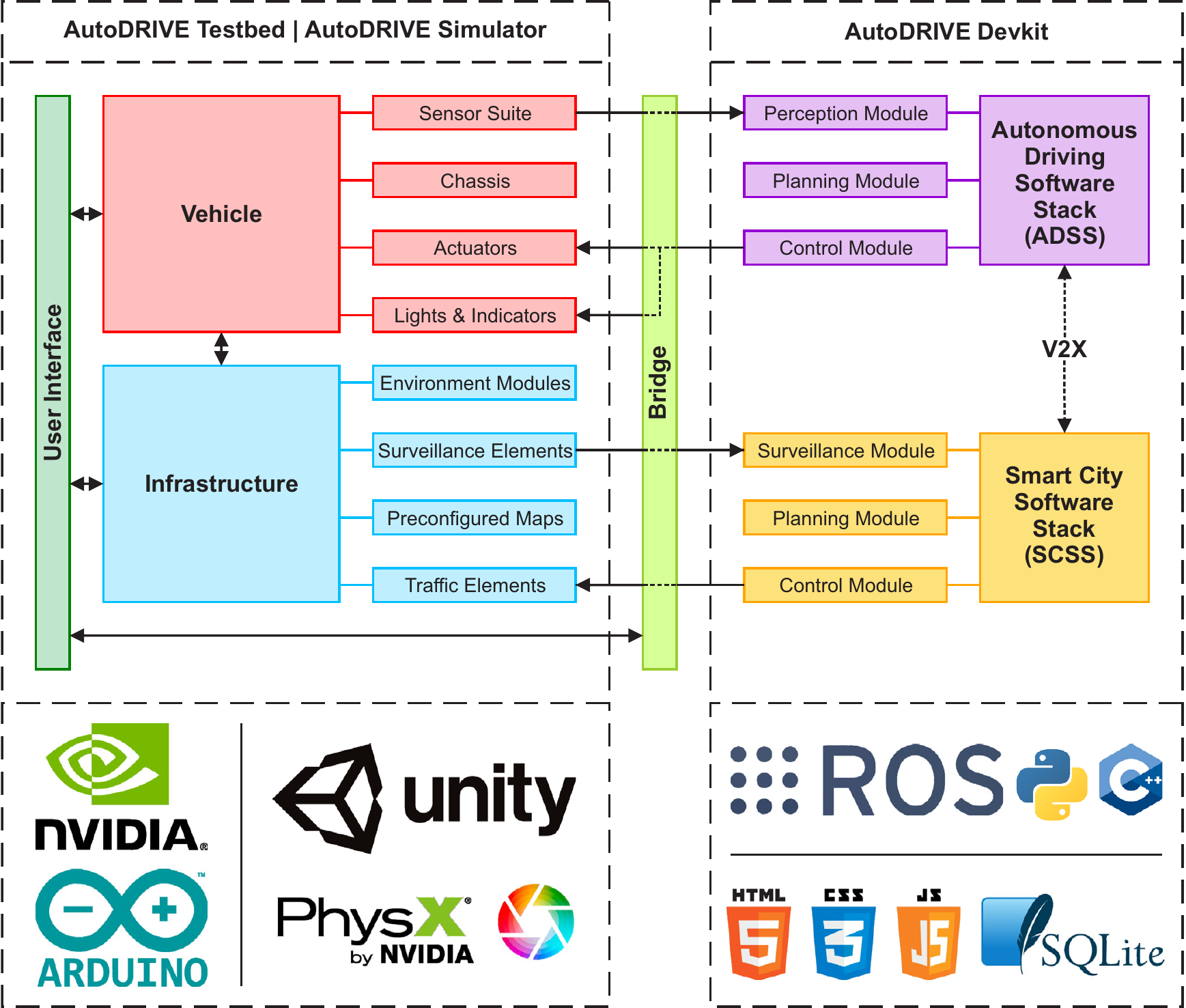}
\caption{High-level overview of the AutoDRIVE Ecosystem, depicting the key modules and their interactions among each other within the three platforms of the ecosystem, viz. AutoDRIVE Testbed, AutoDRIVE Simulator and AutoDRIVE Devkit.}
\label{fig1}
\end{figure}

Advancing the field of connected autonomous vehicles (CAVs) \cite{AD2020} requires scientific and technological research in conjunction with comprehensive education of methods and tools to overcome existing challenges and prepare the next generation of practitioners.

Inasmuch as meaningful verification and validation (V\&V) efforts demand end-to-end stress testing across scales at component/sub-system/system levels, there is a great incentive for exploring the creation and exploitation of varying grades of virtual (simulation-based) and physical (hardware-in-the-loop) testing platforms to alleviate the monetary, spatial, temporal and safety constraints associated with rapid-prototyping of CAV solutions. In a research setting, such platforms can accelerate the process of designing experiments, recording datasets as well as re-iteratively prototyping and validating autonomy solutions. In an educational setting, such platforms can aid in designing interactive demonstrations, hands-on assignments, projects, and competitions.

However, existing platforms for this purpose are observed to limit the throughput of developing and validating connected autonomy solutions. Firstly, most of these platforms lack the integrity required to promote hardware-software co-development; some only offer software simulation tools (e.g., \cite{CARLA2017, LGSVLSimulator2020, AirSim2018}), while others only provide scaled physical vehicles (e.g., \cite{MIT-Racecar2017, AutoRally2021, F1TENTH2019, MuSHR2019}) to test autonomy algorithms. Such isolated platforms not only decelerate the prototyping phase due to compatibility issues, but also adversely affect the validation phase involving simulation to real-world (sim2real) deployments. Secondly, most of these platforms focus specifically on vehicles rather than a holistic intelligent transportation ecosystem involving infrastructure, traffic elements and peer agents, which limits their applications. Thirdly, some of these platforms (e.g., \cite{HyphaROS-Racecar2021, DonkeyCar2021}) are domain specific with limited sensing modalities, stringent design requirements and/or fixed development frameworks; some (e.g. \cite{OCRA2021}) even lack a high-level computation unit and are merely teleoperated from a remote server to execute the intended mission.

\begin{table*}[t]
\centering
\caption{Comparative analysis of scaled platforms/ecosystems for autonomy research and education}
\label{tab1}
\resizebox{\textwidth}{!}{%
    \begin{tabular}{
            >{\columncolor[HTML]{C0C0C0}}l 
            >{\columncolor[HTML]{C6EFCE}}l 
            >{\columncolor[HTML]{FFEB9C}}c 
            >{\columncolor[HTML]{C6EFCE}}c l
            >{\columncolor[HTML]{FFC7CE}}c 
            >{\columncolor[HTML]{FFC7CE}}c c
            >{\columncolor[HTML]{FFC7CE}}c 
            >{\columncolor[HTML]{C6EFCE}}c 
            >{\columncolor[HTML]{C6EFCE}}c 
            >{\columncolor[HTML]{C6EFCE}}c ll
            >{\columncolor[HTML]{C6EFCE}}c 
            >{\columncolor[HTML]{FFC7CE}}c 
            >{\columncolor[HTML]{FFEB9C}}l cc
            >{\columncolor[HTML]{FFC7CE}}c 
            >{\columncolor[HTML]{FFC7CE}}c 
            >{\columncolor[HTML]{FFC7CE}}c 
            >{\columncolor[HTML]{C6EFCE}}c 
            >{\columncolor[HTML]{FFC7CE}}c
            >{\columncolor[HTML]{FFC7CE}}l }
        \multicolumn{1}{c}{\multirow{-2}{*}{\cellcolor[HTML]{85DFFF}\textbf{\begin{tabular}[c]{@{}l@{}}\\Platform/Ecosystem\end{tabular}}}}                                              & \multicolumn{1}{r}{\rot{90}{\cellcolor[HTML]{85DFFF}}}                                 & \multicolumn{1}{r}{\rot{90}{\cellcolor[HTML]{85DFFF}}}                                         & \multicolumn{1}{r}{\rot{90}{\cellcolor[HTML]{85DFFF}}}                                         & \multicolumn{1}{c}{\multirow{-2}{*}{\cellcolor[HTML]{85DFFF}\textbf{\begin{tabular}[c]{@{}l@{}}\\Cost$^{\mathrm{*}}$\end{tabular}}}}                                 & \multicolumn{7}{c}{\cellcolor[HTML]{85DFFF}\textbf{Sensing Modalities}}                                                                                                                                                                                                                                                                                                                                                                                                                                        & \multicolumn{2}{c}{\cellcolor[HTML]{85DFFF}\textbf{Computational Resources}}                                                                     & \multicolumn{2}{c}{\cellcolor[HTML]{85DFFF}\textbf{\begin{tabular}[c]{@{}c@{}}\\Actuation\\Mechanism\end{tabular}}}                                                                                                                                & \multirow{-2}{*}{\cellcolor[HTML]{85DFFF}\textbf{\begin{tabular}[c]{@{}l@{}}\\Dedicated\\Simulator\end{tabular}}}                                                                                                     & \multicolumn{1}{l}{\cellcolor[HTML]{85DFFF}}                                                                                         & \multicolumn{2}{c}{\cellcolor[HTML]{85DFFF}\textbf{V2X Support}}                                                                            & \multicolumn{5}{c}{\cellcolor[HTML]{85DFFF}\textbf{API Support}}                                                                                                                                                                                                                 \\
        \multicolumn{1}{c}{\cellcolor[HTML]{85DFFF}} & \multicolumn{1}{r}{\rot{90}{\multirow{-2}{*}{\cellcolor[HTML]{85DFFF}\textbf{Scale}}}} & \multicolumn{1}{r}{\rot{90}{\multirow{-2}{*}{\cellcolor[HTML]{85DFFF}\textbf{Open Hardware}}}} & \multicolumn{1}{r}{\rot{90}{\multirow{-2}{*}{\cellcolor[HTML]{85DFFF}\textbf{Open Software}}}} & \multicolumn{1}{c}{\cellcolor[HTML]{85DFFF}} & \multicolumn{1}{l}{\rot{90}{\cellcolor[HTML]{85DFFF}\textbf{\textit{Throttle}}}}        & \multicolumn{1}{l}{\rot{90}{\cellcolor[HTML]{85DFFF}\textbf{\textit{Steering}}}}        & \multicolumn{1}{l}{\rot{90}{\cellcolor[HTML]{85DFFF}\textbf{\textit{Wheel Encoders}}}}  & \multicolumn{1}{l}{\rot{90}{\cellcolor[HTML]{85DFFF}\textbf{\textit{GPS/IPS}}}}         & \multicolumn{1}{l}{\rot{90}{\cellcolor[HTML]{85DFFF}\textbf{\textit{IMU}}}}             & \multicolumn{1}{l}{\rot{90}{\cellcolor[HTML]{85DFFF}\textbf{\textit{LIDAR}}}}           & \multicolumn{1}{l}{\rot{90}{\cellcolor[HTML]{85DFFF}\textbf{\textit{Camera}}}}          & \multicolumn{1}{c}{\cellcolor[HTML]{85DFFF}\textbf{\textit{High-Level}}}         & \multicolumn{1}{c}{\cellcolor[HTML]{85DFFF}\textbf{\textit{Low-Level}}}         & \multicolumn{1}{l}{\rot{90}{\cellcolor[HTML]{85DFFF}\textbf{\textit{\begin{tabular}[c]{@{}l@{}}Ackermann\\ Steered\end{tabular}}}}} & \multicolumn{1}{l}{\rot{90}{\cellcolor[HTML]{85DFFF}\textbf{\textit{\begin{tabular}[c]{@{}l@{}}Differetial-Drive/\\ Skid-Steered\end{tabular}}}}} & \multicolumn{1}{l}{\cellcolor[HTML]{85DFFF}} & \multicolumn{1}{r}{\rot{90}{\multirow{-2}{*}{\cellcolor[HTML]{85DFFF}\textbf{\begin{tabular}[c]{@{}l@{}}Multi-Agent\\ Support\end{tabular}}}}} & \cellcolor[HTML]{85DFFF}\textbf{\textit{V2V}}                                 & \cellcolor[HTML]{85DFFF}\textbf{\textit{V2I}}                                 & \rot{90}{\cellcolor[HTML]{85DFFF}\textbf{\textit{C++}}}                                 & \rot{90}{\cellcolor[HTML]{85DFFF}\textbf{\textit{Python}}}                              & \rot{90}{\cellcolor[HTML]{85DFFF}\textbf{\textit{ROS}}}                                 & \rot{90}{\cellcolor[HTML]{85DFFF}\textbf{\textit{MATLAB/Simulink}}} & \rot{90}{\cellcolor[HTML]{85DFFF}\textbf{\textit{Webapp}}} \\
        AutoDRIVE                                                                                 & {\color[HTML]{006100} 1:14}                                                  & \cellcolor[HTML]{C6EFCE}{\color[HTML]{006100} \cmark}                 & {\color[HTML]{006100} \cmark}                                         & \cellcolor[HTML]{C6EFCE}{\color[HTML]{006100} \$450}                        & \cellcolor[HTML]{C6EFCE}{\color[HTML]{006100} \cmark} & \cellcolor[HTML]{C6EFCE}{\color[HTML]{006100} \cmark} & \cellcolor[HTML]{C6EFCE}{\color[HTML]{006100} \cmark} & \cellcolor[HTML]{C6EFCE}{\color[HTML]{006100} \cmark} & {\color[HTML]{006100} \cmark}                         & {\color[HTML]{006100} \cmark}                         & {\color[HTML]{006100} \cmark}                         & \cellcolor[HTML]{C6EFCE}{\color[HTML]{006100} Jetson Nano}              & \cellcolor[HTML]{C6EFCE}{\color[HTML]{006100} Arduino Nano}            & {\color[HTML]{006100} \cmark}                                                                     & \cellcolor[HTML]{FFEB9C}{\color[HTML]{9C5700} \smark}                                                           & \multicolumn{1}{c}{\cellcolor[HTML]{C6EFCE}{\color[HTML]{006100} \cmark}}                                                         & \cellcolor[HTML]{C6EFCE}{\color[HTML]{006100} \cmark}                                                                 & \cellcolor[HTML]{C6EFCE}{\color[HTML]{006100} \cmark} & \cellcolor[HTML]{C6EFCE}{\color[HTML]{006100} \cmark} & \cellcolor[HTML]{C6EFCE}{\color[HTML]{006100} \cmark} & \cellcolor[HTML]{C6EFCE}{\color[HTML]{006100} \cmark} & {\color[HTML]{006100} \cmark}                         & \cellcolor[HTML]{FFEB9C}{\color[HTML]{9C5700} \smark} & \cellcolor[HTML]{C6EFCE}{\color[HTML]{006100} \cmark}       \\
        MIT Racecar                                                                               & {\color[HTML]{006100} 1:10}                                                  & {\color[HTML]{9C5700} \smark}                                         & {\color[HTML]{006100} \cmark}                                         & \cellcolor[HTML]{FFEB9C}{\color[HTML]{9C5700} \$2,600}                      & {\color[HTML]{9C0006} \xmark}                         & {\color[HTML]{9C0006} \xmark}                         & \cellcolor[HTML]{FFC7CE}{\color[HTML]{9C0006} \xmark} & {\color[HTML]{9C0006} \xmark}                         & {\color[HTML]{006100} \cmark}                         & {\color[HTML]{006100} \cmark}                         & {\color[HTML]{006100} \cmark}                         & \cellcolor[HTML]{C6EFCE}{\color[HTML]{006100} Jetson TX2}               & \cellcolor[HTML]{FFEB9C}{\color[HTML]{9C5700} VESC}                    & {\color[HTML]{006100} \cmark}                                                                     & {\color[HTML]{9C0006} \xmark}                                                                                   & {\color[HTML]{9C5700} Gazebo}                                                                                                                    & \cellcolor[HTML]{FFEB9C}{\color[HTML]{9C5700} \smark}                                                                 & \cellcolor[HTML]{FFEB9C}{\color[HTML]{9C5700} \smark} & {\color[HTML]{9C0006} \xmark}                         & {\color[HTML]{9C0006} \xmark}                         & {\color[HTML]{9C0006} \xmark}                         & {\color[HTML]{006100} \cmark}                         & \cellcolor[HTML]{FFC7CE}{\color[HTML]{9C0006} \xmark} & \cellcolor[HTML]{FFC7CE}{\color[HTML]{9C0006} \xmark}       \\
        AutoRally                                                                                 & \cellcolor[HTML]{FFEB9C}{\color[HTML]{9C5700} 1:5}                           & {\color[HTML]{9C5700} \smark}                                         & {\color[HTML]{006100} \cmark}                                         & \cellcolor[HTML]{FFC7CE}{\color[HTML]{9C0006} \$23,300}                     & {\color[HTML]{9C0006} \xmark}                         & {\color[HTML]{9C0006} \xmark}                         & \cellcolor[HTML]{C6EFCE}{\color[HTML]{006100} \cmark} & \cellcolor[HTML]{C6EFCE}{\color[HTML]{006100} \cmark} & {\color[HTML]{006100} \cmark}                         & {\color[HTML]{006100} \cmark}                         & {\color[HTML]{006100} \cmark}                         & \cellcolor[HTML]{FFEB9C}{\color[HTML]{9C5700} Custom}                   & \cellcolor[HTML]{C6EFCE}{\color[HTML]{006100} Teensy LC/Arduino Micro} & {\color[HTML]{006100} \cmark}                                                                     & {\color[HTML]{9C0006} \xmark}                                                                                   & {\color[HTML]{9C5700} Gazebo}                                                                                                                    & \cellcolor[HTML]{FFEB9C}{\color[HTML]{9C5700} \smark}                                                                 & \cellcolor[HTML]{FFEB9C}{\color[HTML]{9C5700} \smark} & {\color[HTML]{9C0006} \xmark}                         & {\color[HTML]{9C0006} \xmark}                         & {\color[HTML]{9C0006} \xmark}                         & {\color[HTML]{006100} \cmark}                         & \cellcolor[HTML]{FFC7CE}{\color[HTML]{9C0006} \xmark} & \cellcolor[HTML]{FFC7CE}{\color[HTML]{9C0006} \xmark}        \\
        F1TENTH                                                                                   & {\color[HTML]{006100} 1:10}                                                  & {\color[HTML]{9C5700} \smark}                                         & {\color[HTML]{006100} \cmark}                                         & \cellcolor[HTML]{FFEB9C}{\color[HTML]{9C5700} \$3,260}                      & {\color[HTML]{9C0006} \xmark}                         & {\color[HTML]{9C0006} \xmark}                         & \cellcolor[HTML]{FFC7CE}{\color[HTML]{9C0006} \xmark} & {\color[HTML]{9C0006} \xmark}                         & \cellcolor[HTML]{FFC7CE}{\color[HTML]{9C0006} \xmark} & {\color[HTML]{006100} \cmark}                         & \cellcolor[HTML]{FFC7CE}{\color[HTML]{9C0006} \xmark} & \cellcolor[HTML]{C6EFCE}{\color[HTML]{006100} Jetson TX2}               & \cellcolor[HTML]{FFEB9C}{\color[HTML]{9C5700} VESC 6MkV}               & {\color[HTML]{006100} \cmark}                                                                     & {\color[HTML]{9C0006} \xmark}                                                                                   & {\color[HTML]{9C5700} RViz/Gazebo}                                                                                                               & \cellcolor[HTML]{C6EFCE}{\color[HTML]{006100} \cmark}                                                                 & \cellcolor[HTML]{C6EFCE}{\color[HTML]{006100} \cmark} & {\color[HTML]{9C0006} \xmark}                         & {\color[HTML]{9C0006} \xmark}                         & {\color[HTML]{9C0006} \xmark}                         & {\color[HTML]{006100} \cmark}                         & \cellcolor[HTML]{FFC7CE}{\color[HTML]{9C0006} \xmark} & \cellcolor[HTML]{FFC7CE}{\color[HTML]{9C0006} \xmark}        \\
        DSV                                                                                       & {\color[HTML]{006100} 1:10}                                                  & {\color[HTML]{9C5700} \smark}                                         & {\color[HTML]{006100} \cmark}                                         & \cellcolor[HTML]{FFEB9C}{\color[HTML]{9C5700} \$1,000}                      & {\color[HTML]{9C0006} \xmark}                         & {\color[HTML]{9C0006} \xmark}                         & \cellcolor[HTML]{C6EFCE}{\color[HTML]{006100} \cmark} & {\color[HTML]{9C0006} \xmark}                         & {\color[HTML]{006100} \cmark}                         & {\color[HTML]{006100} \cmark}                         & {\color[HTML]{006100} \cmark}                         & \cellcolor[HTML]{FFEB9C}{\color[HTML]{9C5700} ODROID-XU4}               & \cellcolor[HTML]{C6EFCE}{\color[HTML]{006100} Arduino (Mega + Uno)}    & {\color[HTML]{006100} \cmark}                                                                     & {\color[HTML]{9C0006} \xmark}                                                                                   & \multicolumn{1}{c}{\cellcolor[HTML]{FFC7CE}{\color[HTML]{9C0006} \xmark}}                                                         & \cellcolor[HTML]{FFC7CE}{\color[HTML]{9C0006} \xmark}                                                                 & \cellcolor[HTML]{FFC7CE}{\color[HTML]{9C0006} \xmark} & {\color[HTML]{9C0006} \xmark}                         & {\color[HTML]{9C0006} \xmark}                         & {\color[HTML]{9C0006} \xmark}                         & {\color[HTML]{006100} \cmark}                         & \cellcolor[HTML]{FFC7CE}{\color[HTML]{9C0006} \xmark} & \cellcolor[HTML]{FFC7CE}{\color[HTML]{9C0006} \xmark}       \\
        MuSHR                                                                                     & {\color[HTML]{006100} 1:10}                                                  & {\color[HTML]{9C5700} \smark}                                         & {\color[HTML]{006100} \cmark}                                         & \cellcolor[HTML]{C6EFCE}{\color[HTML]{006100} \$930}                        & {\color[HTML]{9C0006} \xmark}                         & {\color[HTML]{9C0006} \xmark}                         & \cellcolor[HTML]{FFC7CE}{\color[HTML]{9C0006} \xmark} & {\color[HTML]{9C0006} \xmark}                         & \cellcolor[HTML]{FFC7CE}{\color[HTML]{9C0006} \xmark} & \cellcolor[HTML]{FFC7CE}{\color[HTML]{9C0006} \xmark} & {\color[HTML]{006100} \cmark}                         & \cellcolor[HTML]{C6EFCE}{\color[HTML]{006100} Jetson Nano}              & \cellcolor[HTML]{FFEB9C}{\color[HTML]{9C5700} Turnigy SK8-ESC}         & {\color[HTML]{006100} \cmark}                                                                     & {\color[HTML]{9C0006} \xmark}                                                                                   & {\color[HTML]{9C5700} RViz}                                                                                                                      & \cellcolor[HTML]{C6EFCE}{\color[HTML]{006100} \cmark}                                                                 & \cellcolor[HTML]{C6EFCE}{\color[HTML]{006100} \cmark} & {\color[HTML]{9C0006} \xmark}                         & {\color[HTML]{9C0006} \xmark}                         & {\color[HTML]{9C0006} \xmark}                         & {\color[HTML]{006100} \cmark}                         & \cellcolor[HTML]{FFC7CE}{\color[HTML]{9C0006} \xmark} & \cellcolor[HTML]{FFC7CE}{\color[HTML]{9C0006} \xmark}       \\
        HyphaROS RaceCar                                                                          & {\color[HTML]{006100} 1:10}                                                  & {\color[HTML]{9C5700} \smark}                                         & {\color[HTML]{006100} \cmark}                                         & \cellcolor[HTML]{C6EFCE}{\color[HTML]{006100} \$600}                        & {\color[HTML]{9C0006} \xmark}                         & {\color[HTML]{9C0006} \xmark}                         & \cellcolor[HTML]{FFC7CE}{\color[HTML]{9C0006} \xmark} & {\color[HTML]{9C0006} \xmark}                         & {\color[HTML]{006100} \cmark}                         & {\color[HTML]{006100} \cmark}                         & \cellcolor[HTML]{FFC7CE}{\color[HTML]{9C0006} \xmark} & \cellcolor[HTML]{FFEB9C}{\color[HTML]{9C5700} ODROID-XU4}               & \cellcolor[HTML]{FFEB9C}{\color[HTML]{9C5700} RC ESC TBLE-02S}         & {\color[HTML]{006100} \cmark}                                                                     & {\color[HTML]{9C0006} \xmark}                                                                                   & \multicolumn{1}{c}{\cellcolor[HTML]{FFC7CE}{\color[HTML]{9C0006} \xmark}}                                                         & \cellcolor[HTML]{FFC7CE}{\color[HTML]{9C0006} \xmark}                                                                 & \cellcolor[HTML]{FFC7CE}{\color[HTML]{9C0006} \xmark} & {\color[HTML]{9C0006} \xmark}                         & {\color[HTML]{9C0006} \xmark}                         & {\color[HTML]{9C0006} \xmark}                         & {\color[HTML]{006100} \cmark}                         & \cellcolor[HTML]{FFC7CE}{\color[HTML]{9C0006} \xmark} &
        \cellcolor[HTML]{FFC7CE}{\color[HTML]{9C0006} \xmark}       \\
        Donkey Car                                                                                & {\color[HTML]{006100} 1:16}                                                  & {\color[HTML]{9C5700} \smark}                                         & {\color[HTML]{006100} \cmark}                                         & \cellcolor[HTML]{C6EFCE}{\color[HTML]{006100} \$370}                        & {\color[HTML]{9C0006} \xmark}                         & {\color[HTML]{9C0006} \xmark}                         & \cellcolor[HTML]{FFC7CE}{\color[HTML]{9C0006} \xmark} & {\color[HTML]{9C0006} \xmark}                         & \cellcolor[HTML]{FFC7CE}{\color[HTML]{9C0006} \xmark} & \cellcolor[HTML]{FFC7CE}{\color[HTML]{9C0006} \xmark} & {\color[HTML]{006100} \cmark}                         & \cellcolor[HTML]{FFEB9C}{\color[HTML]{9C5700} Raspberry Pi}             & \cellcolor[HTML]{FFEB9C}{\color[HTML]{9C5700} ESC}                     & {\color[HTML]{006100} \cmark}                                                                     & {\color[HTML]{9C0006} \xmark}                                                                                   & {\color[HTML]{9C5700} Gym}                                                                                                                       & \cellcolor[HTML]{FFC7CE}{\color[HTML]{9C0006} \xmark}                                                                 & \cellcolor[HTML]{FFC7CE}{\color[HTML]{9C0006} \xmark} & {\color[HTML]{9C0006} \xmark}                         & {\color[HTML]{9C0006} \xmark}                         & \cellcolor[HTML]{C6EFCE}{\color[HTML]{006100} \cmark} & \cellcolor[HTML]{FFC7CE}{\color[HTML]{9C0006} \xmark} & \cellcolor[HTML]{FFC7CE}{\color[HTML]{9C0006} \xmark} &
        \cellcolor[HTML]{FFC7CE}{\color[HTML]{9C0006} \xmark}       \\
        BARC                                                                                      & {\color[HTML]{006100} 1:10}                                                  & {\color[HTML]{9C5700} \smark}                                         & {\color[HTML]{006100} \cmark}                                         & \cellcolor[HTML]{FFEB9C}{\color[HTML]{9C5700} \$1,030}                      & {\color[HTML]{9C0006} \xmark}                         & {\color[HTML]{9C0006} \xmark}                         & \cellcolor[HTML]{C6EFCE}{\color[HTML]{006100} \cmark} & {\color[HTML]{9C0006} \xmark}                         & {\color[HTML]{006100} \cmark}                         & \cellcolor[HTML]{FFC7CE}{\color[HTML]{9C0006} \xmark} & {\color[HTML]{006100} \cmark}                         & \cellcolor[HTML]{FFEB9C}{\color[HTML]{9C5700} ODROID-XU4}               & \cellcolor[HTML]{C6EFCE}{\color[HTML]{006100} Arduino Nano}            & {\color[HTML]{006100} \cmark}                                                                     & {\color[HTML]{9C0006} \xmark}                                                                                   & \multicolumn{1}{c}{\cellcolor[HTML]{FFC7CE}{\color[HTML]{9C0006} \xmark}}                                                         & \cellcolor[HTML]{FFC7CE}{\color[HTML]{9C0006} \xmark}                                                                 & \cellcolor[HTML]{FFC7CE}{\color[HTML]{9C0006} \xmark} & {\color[HTML]{9C0006} \xmark}                         & {\color[HTML]{9C0006} \xmark}                         & {\color[HTML]{9C0006} \xmark}                         & {\color[HTML]{006100} \cmark}                         & \cellcolor[HTML]{FFC7CE}{\color[HTML]{9C0006} \xmark} & \cellcolor[HTML]{FFC7CE}{\color[HTML]{9C0006} \xmark}       \\
        OCRA                                                                                & {\color[HTML]{006100} 1:43}                           & {\color[HTML]{9C5700} \smark}                 & {\color[HTML]{006100} \cmark}                                         & \cellcolor[HTML]{C6EFCE}{\color[HTML]{006100} \$960}                        & {\color[HTML]{9C0006} \xmark}                         & {\color[HTML]{9C0006} \xmark}                         & \cellcolor[HTML]{FFC7CE}{\color[HTML]{9C0006} \xmark} & \cellcolor[HTML]{FFC7CE}{\color[HTML]{9C0006} \xmark} & {\color[HTML]{006100} \cmark}                         & \cellcolor[HTML]{FFC7CE}{\color[HTML]{9C0006} \xmark} & \cellcolor[HTML]{FFC7CE}{\color[HTML]{9C0006} \xmark} & \cellcolor[HTML]{FFC7CE}{\color[HTML]{9C0006} None}             & \cellcolor[HTML]{C6EFCE}{\color[HTML]{006100} ARM Cortex M4 $\mu$C}                  & \cellcolor[HTML]{C6EFCE}{\color[HTML]{006100} \cmark}                                             & \cellcolor[HTML]{FFC7CE}{\color[HTML]{9C0006} \xmark}                                                           & \multicolumn{1}{c}{\cellcolor[HTML]{FFC7CE}{\color[HTML]{9C0006} \xmark}}                                                                                                                    & \cellcolor[HTML]{C6EFCE}{\color[HTML]{006100} \cmark}  & \cellcolor[HTML]{FFC7CE}{\color[HTML]{9C0006} \xmark}    & \cellcolor[HTML]{FFC7CE}{\color[HTML]{9C0006} \xmark}    & \cellcolor[HTML]{C6EFCE}{\color[HTML]{006100} \cmark}    & \cellcolor[HTML]{FFC7CE}{\color[HTML]{9C0006} \xmark}    & \cellcolor[HTML]{FFC7CE}{\color[HTML]{9C0006} \xmark}    & \cellcolor[HTML]{C6EFCE}{\color[HTML]{006100} \cmark}    &
        \cellcolor[HTML]{FFC7CE}{\color[HTML]{9C0006} \xmark} \\
        QCar                                                                                      & {\color[HTML]{006100} 1:10}                                                  & \cellcolor[HTML]{FFC7CE}{\color[HTML]{9C0006} \xmark}                 & \cellcolor[HTML]{FFC7CE}{\color[HTML]{9C0006} \xmark}                 & \cellcolor[HTML]{FFC7CE}{\color[HTML]{9C0006} \$20,000}                     & {\color[HTML]{9C0006} \xmark}                         & {\color[HTML]{9C0006} \xmark}                         & \cellcolor[HTML]{C6EFCE}{\color[HTML]{006100} \cmark} & {\color[HTML]{9C0006} \xmark}                         & {\color[HTML]{006100} \cmark}                         & {\color[HTML]{006100} \cmark}                         & {\color[HTML]{006100} \cmark}                         & \cellcolor[HTML]{C6EFCE}{\color[HTML]{006100} Jetson TX2}               & \cellcolor[HTML]{FFC7CE}{\color[HTML]{9C0006} Proprietary}             & {\color[HTML]{006100} \cmark}                                                                     & {\color[HTML]{9C0006} \xmark}                                                                                   & {\color[HTML]{9C5700} Simulink}                                                                                                                  & \cellcolor[HTML]{C6EFCE}{\color[HTML]{006100} \cmark}                                                                 & \cellcolor[HTML]{C6EFCE}{\color[HTML]{006100} \cmark} & {\color[HTML]{9C0006} \xmark}                         & \cellcolor[HTML]{FFEB9C}{\color[HTML]{9C5700} \smark} & \cellcolor[HTML]{FFEB9C}{\color[HTML]{9C5700} \smark} & \cellcolor[HTML]{FFEB9C}{\color[HTML]{9C5700} \smark} & \cellcolor[HTML]{C6EFCE}{\color[HTML]{006100} \cmark} &     
        \cellcolor[HTML]{FFC7CE}{\color[HTML]{9C0006} \xmark}       \\
        AWS DeepRacer                                                                             & {\color[HTML]{006100} 1:18}                                                  & \cellcolor[HTML]{FFC7CE}{\color[HTML]{9C0006} \xmark}                 & \cellcolor[HTML]{FFC7CE}{\color[HTML]{9C0006} \xmark}                 & \cellcolor[HTML]{C6EFCE}{\color[HTML]{006100} \$400}                        & {\color[HTML]{9C0006} \xmark}                         & {\color[HTML]{9C0006} \xmark}                         & \cellcolor[HTML]{FFC7CE}{\color[HTML]{9C0006} \xmark} & {\color[HTML]{9C0006} \xmark}                         & {\color[HTML]{006100} \cmark}                         & \cellcolor[HTML]{FFEB9C}{\color[HTML]{9C5700} \smark} & {\color[HTML]{006100} \cmark}                         & \cellcolor[HTML]{FFC7CE}{\color[HTML]{9C0006} Proprietary}              & \cellcolor[HTML]{FFC7CE}{\color[HTML]{9C0006} Proprietary}             & {\color[HTML]{006100} \cmark}                                                                     & {\color[HTML]{9C0006} \xmark}                                                                                   & {\color[HTML]{9C5700} Gym}                                                                                                                       & \cellcolor[HTML]{FFC7CE}{\color[HTML]{9C0006} \xmark}                                                                 & \cellcolor[HTML]{FFC7CE}{\color[HTML]{9C0006} \xmark} & {\color[HTML]{9C0006} \xmark}                         & {\color[HTML]{9C0006} \xmark}                         & {\color[HTML]{9C0006} \xmark}                         & \cellcolor[HTML]{FFC7CE}{\color[HTML]{9C0006} \xmark} &
        \cellcolor[HTML]{FFC7CE}{\color[HTML]{9C0006} \xmark} & \cellcolor[HTML]{C6EFCE}{\color[HTML]{006100} \cmark}       \\
        Duckietown                                                                                & \cellcolor[HTML]{FFC7CE}{\color[HTML]{9C0006} N/A}                           & \cellcolor[HTML]{C6EFCE}{\color[HTML]{006100} \cmark}                 & {\color[HTML]{006100} \cmark}                                         & \cellcolor[HTML]{C6EFCE}{\color[HTML]{006100} \$450}                        & {\color[HTML]{9C0006} \xmark}                         & {\color[HTML]{9C0006} \xmark}                         & \cellcolor[HTML]{FFEB9C}{\color[HTML]{9C5700} \smark} & {\color[HTML]{9C0006} \xmark}                         & \cellcolor[HTML]{FFEB9C}{\color[HTML]{9C5700} \smark} & \cellcolor[HTML]{FFC7CE}{\color[HTML]{9C0006} \xmark} & {\color[HTML]{006100} \cmark}                         & \cellcolor[HTML]{FFEB9C}{\color[HTML]{9C5700} Raspberry Pi/Jetson Nano} & \cellcolor[HTML]{FFC7CE}{\color[HTML]{9C0006} None}                    & \cellcolor[HTML]{FFC7CE}{\color[HTML]{9C0006} \xmark}                                             & \cellcolor[HTML]{C6EFCE}{\color[HTML]{006100} \cmark}                                                           & {\color[HTML]{9C5700} Gym}                                                                                                                       & \cellcolor[HTML]{C6EFCE}{\color[HTML]{006100} \cmark}                                                                 & \cellcolor[HTML]{FFC7CE}{\color[HTML]{9C0006} \xmark} & \cellcolor[HTML]{FFEB9C}{\color[HTML]{9C5700} \smark} & {\color[HTML]{9C0006} \xmark}                         & {\color[HTML]{9C0006} \xmark}                         & {\color[HTML]{006100} \cmark}                         & \cellcolor[HTML]{FFC7CE}{\color[HTML]{9C0006} \xmark} & \cellcolor[HTML]{FFC7CE}{\color[HTML]{9C0006} \xmark}       \\
        TurtleBot3                                                                                & \cellcolor[HTML]{FFC7CE}{\color[HTML]{9C0006} N/A}                           & \cellcolor[HTML]{C6EFCE}{\color[HTML]{006100} \cmark}                 & {\color[HTML]{006100} \cmark}                                         & \cellcolor[HTML]{C6EFCE}{\color[HTML]{006100} \$590}                        & {\color[HTML]{9C0006} \xmark}                         & {\color[HTML]{9C0006} \xmark}                         & \cellcolor[HTML]{C6EFCE}{\color[HTML]{006100} \cmark} & {\color[HTML]{9C0006} \xmark}                         & {\color[HTML]{006100} \cmark}                         & {\color[HTML]{006100} \cmark}                         & \cellcolor[HTML]{FFEB9C}{\color[HTML]{9C5700} \smark} & \cellcolor[HTML]{FFEB9C}{\color[HTML]{9C5700} Raspberry Pi}             & \cellcolor[HTML]{C6EFCE}{\color[HTML]{006100} OpenCR}                  & \cellcolor[HTML]{FFC7CE}{\color[HTML]{9C0006} \xmark}                                             & \cellcolor[HTML]{C6EFCE}{\color[HTML]{006100} \cmark}                                                           & {\color[HTML]{9C5700} Gazebo}                                                                                                                    & \cellcolor[HTML]{FFEB9C}{\color[HTML]{9C5700} \smark}                                                                 & \cellcolor[HTML]{FFEB9C}{\color[HTML]{9C5700} \smark} & {\color[HTML]{9C0006} \xmark}                         & {\color[HTML]{9C0006} \xmark}                         & {\color[HTML]{9C0006} \xmark}                         & {\color[HTML]{006100} \cmark}                         & \cellcolor[HTML]{FFC7CE}{\color[HTML]{9C0006} \xmark} &
        \cellcolor[HTML]{FFC7CE}{\color[HTML]{9C0006} \xmark}       \\
        Pheeno                                                                                & \cellcolor[HTML]{FFC7CE}{\color[HTML]{9C0006} N/A}                           & \cellcolor[HTML]{C6EFCE}{\color[HTML]{006100} \cmark}                 & {\color[HTML]{006100} \cmark}                                         & \cellcolor[HTML]{C6EFCE}{\color[HTML]{006100} \$350}                        & {\color[HTML]{9C0006} \xmark}                         & {\color[HTML]{9C0006} \xmark}                         & \cellcolor[HTML]{C6EFCE}{\color[HTML]{006100} \cmark} & \cellcolor[HTML]{FFC7CE}{\color[HTML]{9C0006} \xmark} & {\color[HTML]{006100} \cmark}                         & \cellcolor[HTML]{FFC7CE}{\color[HTML]{9C0006} \xmark} & \cellcolor[HTML]{C6EFCE}{\color[HTML]{006100} \cmark} & \cellcolor[HTML]{FFEB9C}{\color[HTML]{9C5700} Raspberry Pi}             & \cellcolor[HTML]{C6EFCE}{\color[HTML]{006100} Arduino Pro Mini}                  & \cellcolor[HTML]{FFC7CE}{\color[HTML]{9C0006} \xmark}                                             & \cellcolor[HTML]{C6EFCE}{\color[HTML]{006100} \cmark}                                                           & \multicolumn{1}{c}{\cellcolor[HTML]{FFC7CE}{\color[HTML]{9C0006} \xmark}}                                                                                                                    & \cellcolor[HTML]{C6EFCE}{\color[HTML]{006100} \cmark}  & \cellcolor[HTML]{C6EFCE}{\color[HTML]{006100} \cmark}    & \cellcolor[HTML]{FFC7CE}{\color[HTML]{9C0006} \xmark}    & \cellcolor[HTML]{FFC7CE}{\color[HTML]{9C0006} \xmark}    & \cellcolor[HTML]{C6EFCE}{\color[HTML]{006100} \cmark}    & \cellcolor[HTML]{FFEB9C}{\color[HTML]{9C5700} \smark}    & \cellcolor[HTML]{FFC7CE}{\color[HTML]{9C0006} \xmark}    &
        \cellcolor[HTML]{FFC7CE}{\color[HTML]{9C0006} \xmark}       \\
        \multicolumn{24}{l}{} \\
        \multicolumn{24}{l}{\color[HTML]{006100} \cmark \color{black} $\!$ indicates complete fulfillment; \color[HTML]{9C5700} \smark \color{black} $\!$ indicates conditional, unsupported or partial fulfillment; and \color[HTML]{9C0006} \xmark \color{black} $\!$ indicates non-fulfillment. $^{\mathrm{*}}$All cost values are ceiled to the nearest \$10.} \\
    \end{tabular}
}
\end{table*}

This work does not propose ``yet another'' research and education platform targeting selective aspects of autonomous driving technology. AutoDRIVE\footnote{\url{https://autodrive-ecosystem.github.io}} (refer Fig. \ref{fig1}) aims to provide a cyber-physical ecosystem that is:
\begin{itemize}
\item \textit{Comprehensive:} The ecosystem offers a scaled car-like vehicle with abundant sensors, which supports single as well as multi-agent algorithms with or without vehicle-to-vehicle (V2V) communication. It also provides a modular infrastructure development kit comprising various environment modules, traffic elements and surveillance elements, which supports internet-of-things (IoT) and vehicle-to-infrastructure (V2I) communication. On the software front, the ecosystem hosts a high-fidelity simulator and supports development of autonomous driving as well as smart city solutions.
\item \textit{Flexible:} The ecosystem offers modular hardware components, a convenient high-fidelity simulator, and an extensive software development support, which enables the end-users to flexibly prototype and validate their autonomy solutions right out-of-the-box. Additionally, the completely open-hardware, open-software architecture of the ecosystem allows users to adapt any of the existing hardware (including design of the vehicle as well as the infrastructure modules) and/or software (including codebase of the development framework as well as the simulator) to better fit their use-cases.
\item \textit{Integrated:} The ecosystem hosts a tightly-coupled trio, comprising AutoDRIVE Devkit (to flexibly develop connected autonomy solutions), AutoDRIVE Simulator (to virtually prototype and test them under a variety of conditions and edge-cases), and AutoDRIVE Testbed (to deploy and validate them in controlled real-world settings). The harmony among these three platforms not only enhances the hardware-software co-development of autonomy solutions, but also helps seamlessly bridge the gap between software simulation and hardware deployment for verification and validation of these safety-critical systems.
\end{itemize}

This work also describes sample use-cases of four emergent applications in the field of CAVs, with each exploiting, and thus exhibiting, distinct features and capabilities of the AutoDRIVE Ecosystem. These include state-of-the-art (SOTA) implementations such as autonomous parking, behavioral cloning and intersection traversal along with a novel implementation of smart city management.


\section{State of the Art}
\label{Section: State of the Art}

\begin{figure*}[t]
\centering
\includegraphics[width=\linewidth]{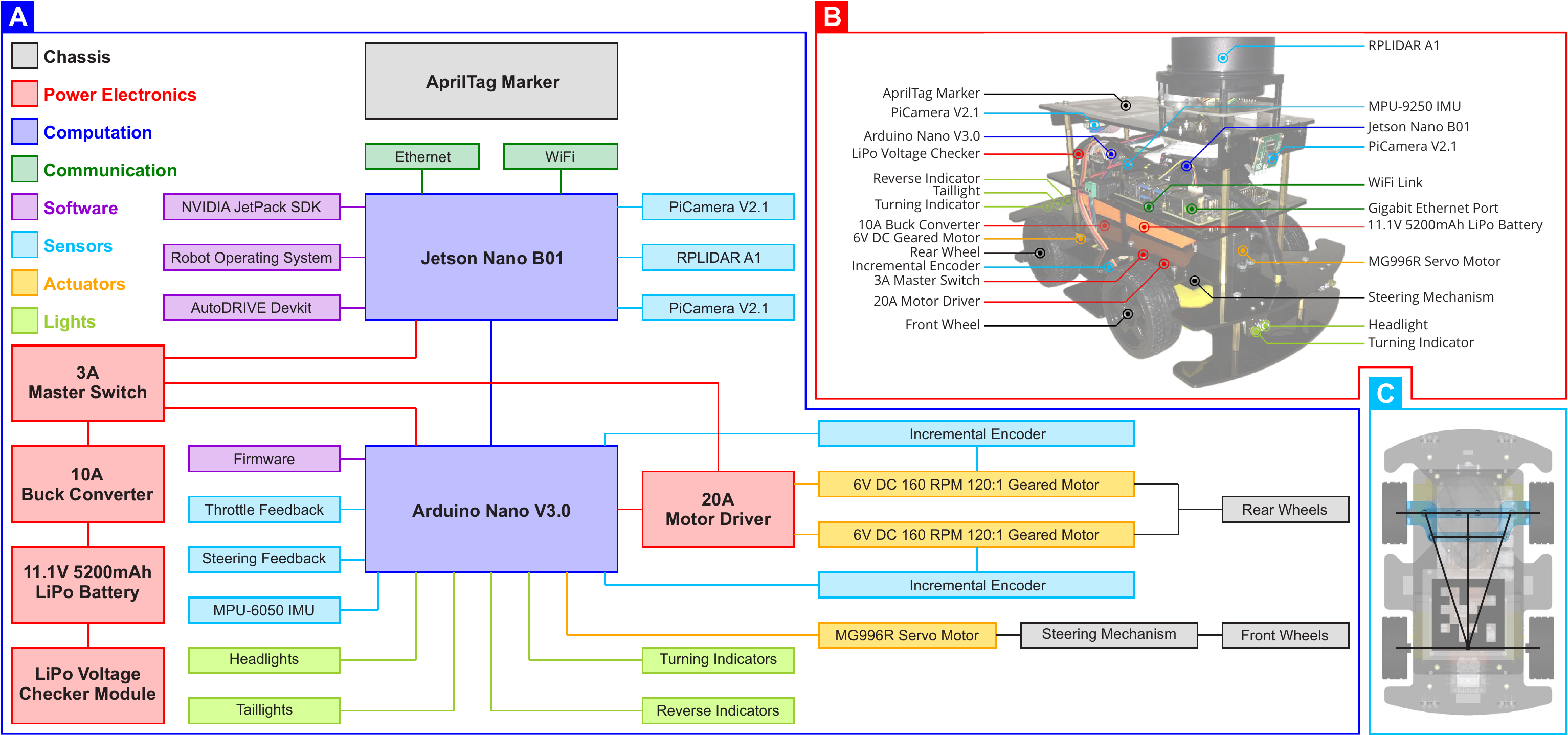}
\caption{Native vehicle (Nigel) of the AutoDRIVE Ecosystem: [A] high-level architecture of the vehicle; [B] various components and sub-systems of the vehicle; [C] open-source chassis of the vehicle adopting rear wheel drive, Ackermann steered actuation mechanism.}
\label{fig2}
\end{figure*}

Deployment of full-scale CAV solutions generally requires extensive verification and validation, which poses several challenges, especially in university settings. The time, expense, resources and knowledge of full-scale testing interplay with infrastructural requirements as well as safety of personnel and property involved, often hindering research and education progress. Consequently, the past decade has witnessed many university-based deployments exploring development of scaled autonomous vehicles. As described in Table \ref{tab1}, such vehicles include the MIT Racecar \cite{MIT-Racecar2017}, AutoRally \cite{AutoRally2021}, F1TENTH \cite{F1TENTH2019}, Multi-agent System for non-Holonomic Racing (MuSHR) \cite{MuSHR2019}, Optimal RC Racing (ORCA) Project \cite{OCRA2021}, Delft Scaled Vehicle (DSV) \cite{DSV2017}, and Berkeley Autonomous Race Car (BARC) \cite{BARC2021}, to name a few. Some of the other community-driven platforms for autonomous driving include HyphaROS RaceCar \cite{HyphaROS-Racecar2021} and Donkey Car \cite{DonkeyCar2021}, both of which are application specific - the former for map-based navigation, and the later for vision-aided imitation learning. Apart from these, commercial products such as QCar \cite{QCar2021} by Quanser and DeepRacer \cite{DeepRacer2021} by Amazon Web Services (AWS) are now surfacing the market. However, the fact that most of these products are expensive and employ some form of proprietary hardware and/or software components restricts their openness and flexibility to the community; not to mention potential issues like warranty-voids and vendor lock-ins. Some of the other scaled platforms for autonomy research and education include Duckietown \cite{Duckietown2017}, TurtleBot3 \cite{Turtlebot2021} and Pheeno \cite{Pheeno2016}. However, the differentially-driven robots proposed by these platforms/ecosystems fail to fully satisfy the community requirements for a kinodynamically constrained car-like vehicle.

In terms of comprehensiveness, some of these platforms lack diverse sensing modalities, some lack adequate computational power, some lack Ackermann steering mechanism, and most lack active or passive infrastructural elements. Only a few satisfy the prominent community requirements, but can prove to be prohibitively expensive for university programs.

In terms of flexibility, most of these platforms, if not all, use commercial-off-the-shelf (COTS) radio-controlled (RC) cars as their base-chassis, which: (a) are quite expensive; (b) may not be available all around the world; (c) limit research on the ``mechatronics engineering'' front, which is equally important for cyber-physical systems such as CAVs. Additionally, most of these platforms only support a specific software framework such as Robot Operating System (ROS) \cite{ROS2009}, which inherently creates a skillset-dependency for the end-users. Furthermore, providing assets and plugins for pre-packaged simulators like Gazebo \cite{Gazebo2004} or OpenAI Gym \cite{OpenAIGym2016} environments offers only so much flexibility to the users in terms of designing and running the simulated scenarios.

In terms of integrity, some of these platforms do not support simulation in any form, some ROS-based ones support kinematic/dynamic simulation using RViz \cite{RViz2021} and/or Gazebo, while others offer task-specific Gym environments for imitation/reinforcement learning; none of which is ideal.


\section{AutoDRIVE Testbed}
\label{Section: AutoDRIVE Testbed}

\begin{figure*}[t]
\centering
\includegraphics[width=\linewidth]{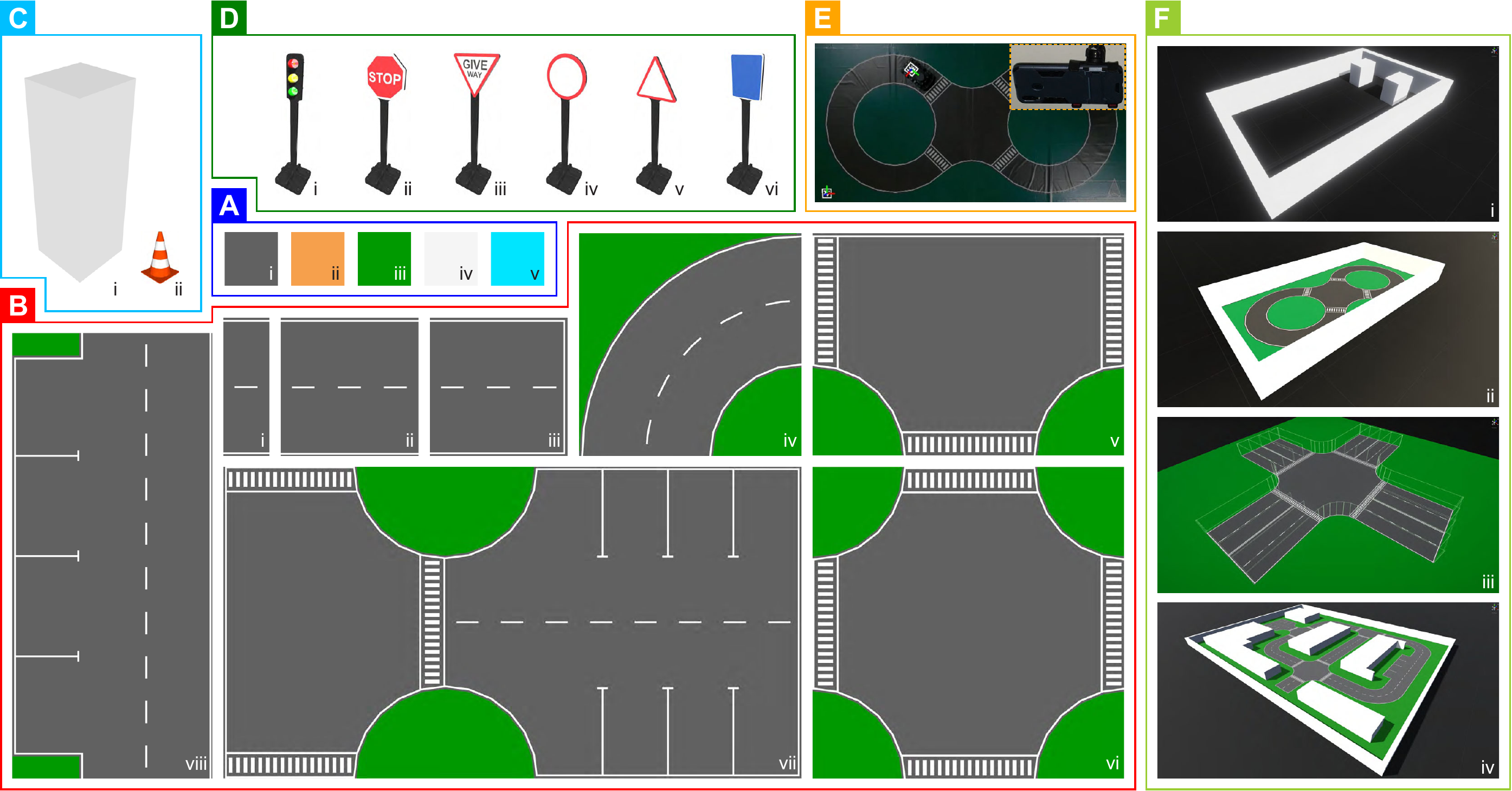}
\caption{Modular and reconfigurable infrastructure development kit of the AutoDRIVE Ecosystem: [A] terrain modules - (i) asphalt, (ii) dirt, (iii) lawn, (iv) snow, (v) water; [B] road kit - (i) road patch, (ii) straight road, (iii) dead-end, (iv) curved road, (v) 3-way intersection, (vi) 4-way intersection, (vii) parking lot, (viii) roadside parking; [C] obstruction modules - (i) construction box, (ii) traffic cone; [D] traffic elements - (i) traffic light, (ii) stop sign, (iii) give way sign, (iv) regulatory sign, (v) cautionary sign, (vi) informatory sign; [E] surveillance elements - vehicle localization using the AutoDRIVE Eye; [F] preconfigured maps - (i) Parking School, (ii) Driving School, (iii) Intersection School, (iv) Tiny Town.}
\label{fig3}
\end{figure*}

AutoDRIVE Testbed is a hardware platform featuring a native scaled vehicle along with a modular and reconfigurable infrastructure development kit for deploying and validating autonomy algorithms in controlled real-world settings. It adopts a completely open-hardware and open-software architecture so as to push the ``systems engineering and integration'' front with regard to CAVs.

\subsection{Vehicle}
\label{Sub-Section: Vehicle}

AutoDRIVE's native vehicle, named Nigel (refer Fig. \hyperref[fig2]{\ref*{fig2}-A} and Fig. \hyperref[fig2]{\ref*{fig2}-B}) offers realistic driving and steering actuation, comprehensive sensor suite, high-performance computational resources, and a standard vehicular lighting system.

\subsubsection{Chassis}
\label{Sub-Sub-Section: Chassis}

Nigel is a 1:14 scale model vehicle comprising four modular platforms, each housing distinct components of the vehicle. It adopts rear wheel drive, Ackermann steered mechanism (refer Fig. \hyperref[fig2]{\ref*{fig2}-C}), thereby resembling an actual car in terms of kinodynamic constraints.

\subsubsection{Power Electronics}
\label{Sub-Sub-Section: Power Electronics}

Nigel is powered using an 11.1V 5200mAh lithium-polymer (LiPo) battery, whose health is monitored by a voltage checker. A 10A rated buck converter steps down the voltage to 5V, which is then routed, via a 3A rated master switch, to all the electrical sub-systems including a 20A rated motor driver module.

\subsubsection{Sensor Suite}
\label{Sub-Sub-Section: Sensor Suite}

Nigel hosts a comprehensive sensor suite comprising throttle and steering sensors (actuator feedbacks), 1920 CPR incremental encoders (wheel rotation/velocity), a 3-axis indoor-positioning system (IPS) using fiducial markers (mm/cm-level accurate small-scale positioning analogous to m-level accurate full-scale GNSS), 9-axis IMU (raw inertial data and calibrated AHRS data using Madgwick/Mahony filter), two 62.2$^\circ$ FOV cameras with 3.04 mm focal length (front and/or rear RGB frames) and a 7-10 Hz, 360$^\circ$ FOV LIDAR with 12 m range and 1$^\circ$ resolution (2D laser scan).

\subsubsection{Computation, Communication and Software}
\label{Sub-Sub-Section: Computation, Communication and Software}

Nigel adopts Jetson Nano Developer Kit - B01 for most of its high-level computation (autonomy algorithms), communication (V2V and V2I) and software installation (JetPack SDK, ROS Melodic and AutoDRIVE Devkit). Additionally, it also hosts Arduino Nano (running the vehicle firmware) for acquiring and filtering raw sensor data and controlling actuators/lights.

\subsubsection{Actuators}
\label{Sub-Sub-Section: Actuators}

Nigel is provided with two 6V 160 RPM rated 120:1 DC geared motors to drive its rear wheels, and a 9.4 kgf.cm servo motor to steer its front wheels; the steering actuator is saturated at $\pm$ 30$^\circ$ w.r.t. zero-steer value. All the actuators are operated at 5V, which translates to a maximum speed of $\sim$130 RPM for driving ($\sim$0.267 m/s @ vehicle) and $\sim$0.19 s/60$^\circ$ for steering ($\sim$0.805 rad/s @ vehicle).

\subsubsection{Lights and Indicators}
\label{Sub-Sub-Section: Lights and Indicators}

Nigel's lighting system comprises of dual-mode headlights, automated taillights, triple-mode turning indicators and automated reverse indicators.

\subsection{Infrastructure}
\label{Sub-Section: Infrastructure}

\begin{figure*}[t]
\centering
\includegraphics[width=\linewidth]{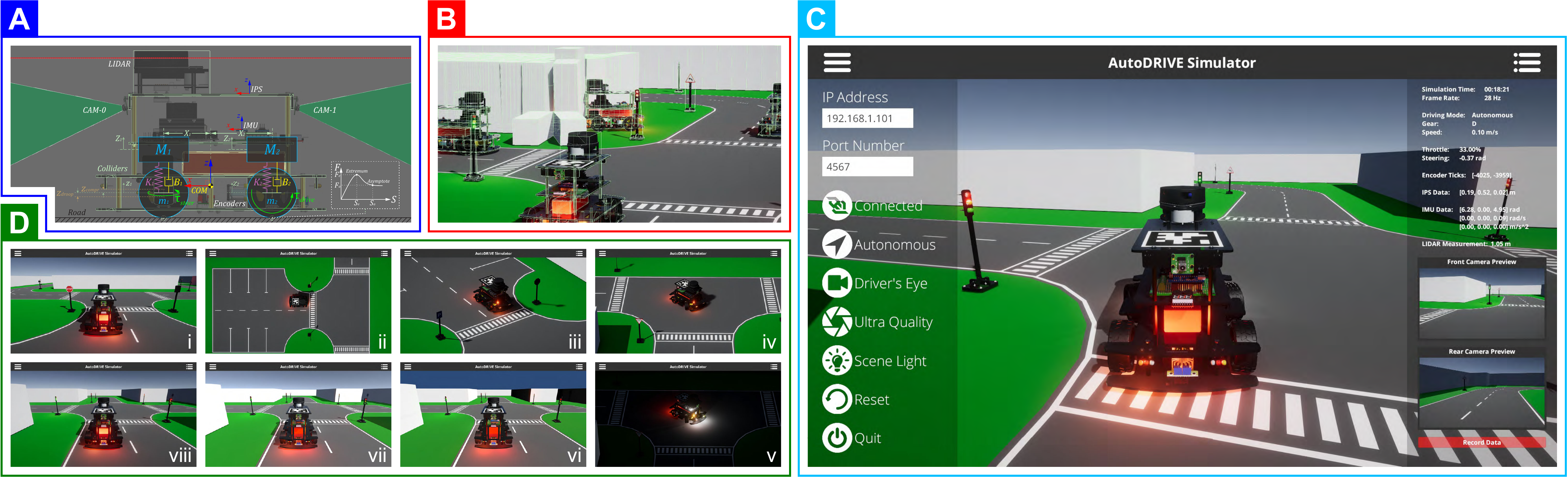}
\caption{Native simulation platform of the AutoDRIVE Ecosystem: [A] simulation of vehicle dynamics, sensors and actuators; [B] simulation of infrastructure dynamics and interaction physics; [C] graphical user interface of the simulator; [D] simulator features - (i) Driver's Eye camera, (ii) Bird's Eye camera, (iii) God's Eye camera, (iv) scene light enabled, (v) scene light disabled, (vi) low-quality graphics, (vii) high-quality graphics, (viii) ultra-quality graphics.}
\label{fig4}
\end{figure*}

AutoDRIVE offers a modular and reconfigurable infrastructure development kit (refer Fig. \ref{fig3}) for rapidly designing and prototyping custom driving scenarios. This kit includes a range of environment modules, traffic elements and surveillance elements, along with several preconfigured maps.

\subsubsection{Environment Modules}
\label{Sub-Sub-Section: Environment Modules}

Environment modules include static layouts and objects meant for rapidly designing custom scenarios. Apart from these, experts may also choose to design scaled real-world or imaginary scenarios using third-party tools, and import them into AutoDRIVE Ecosystem.

\begin{itemize}
\item \textit{Terrain Modules:} These define off-road segments of the environment. AutoDRIVE currently supports 5 terrains with tunable physical properties (refer Fig. \hyperref[fig3]{\ref*{fig3}-A}).
\item \textit{Road Kits:} These enable reconfigurable construction of drivable segments of the environment. AutoDRIVE currently supports 1, 2, 4 and 6 lane road kits, each having 8 different modules (refer Fig. \hyperref[fig3]{\ref*{fig3}-B}).
\item \textit{Obstruction Modules:} These 3D objects define static obstacles within the scene. AutoDRIVE currently supports two such modules (refer Fig. \hyperref[fig3]{\ref*{fig3}-C}).
\end{itemize}

\subsubsection{Traffic Elements}
\label{Sub-Sub-Section: Traffic Elements}

Traffic elements (refer Fig. \hyperref[fig3]{\ref*{fig3}-D}) define traffic laws within a particular driving scenario, thereby governing the traffic flow. AutoDRIVE currently supports modular traffic signs and lights. These modules support IoT and V2I communication technologies, and can be therefore integrated with AutoDRIVE Smart City Manager (SCM).

\subsubsection{Surveillance Elements}
\label{Sub-Sub-Section: Surveillance Elements}

AutoDRIVE features a surveillance element called AutoDRIVE Eye to view the entire scene from a bird's eye view. The said element is also integrated with AutoDRIVE SCM, and, upon calibration of its intrinsic parameters, is capable of estimating vehicle's 2D pose within the map by detecting and tracking the AprilTag markers attached to each of them; this functionality is illustrated in Fig. \hyperref[fig3]{\ref*{fig3}-E} (notice the roof-mounted camera).

\subsubsection{Preconfigured Maps}
\label{Sub-Sub-Section: Preconfigured Maps}

AutoDRIVE currently offers 4 preconfigured maps (refer Fig. \hyperref[fig3]{\ref*{fig3}-F}). Parking School is designed specifically for autonomous parking applications, wherein construction boxes define static obstacles and all the available free-space is drivable. Driving School covers driving over straight roads, curved roads and crossing an intersection. Intersection School is designed specifically for intersection traversal applications, wherein lane bounds play an important role. Finally, Tiny Town is meant to be a comprehensive driving scenario, which covers each and every infrastructure element currently available in AutoDRIVE.

\section{AutoDRIVE Simulator}
\label{Section: AutoDRIVE Simulator}

AutoDRIVE Simulator \cite{AutoDRIVESimulator2021, AutoDRIVESimulatorReport2020} acts as the digital twin of AutoDRIVE Testbed. It is primarily targeted towards virtual prototyping of autonomy solutions, either for variability testing, or as a part of recursive simulation-deployment workflow, but can also be used for synthetic data generation.

\subsection{Vehicle Dynamics Simulation}
\label{Sub-Section: Vehicle Dynamics Simulation}

The vehicle is jointly modelled (refer Fig. \hyperref[fig4]{\ref*{fig4}-A}) as a rigid-body and a collection of sprung masses $^iM$, such that the total mass of the rigid-body is $M=\sum{^iM}$. The rigid-body center of mass $X_{COM} = \frac{\sum{{^iM}*{^iX}}}{\sum{^iM}}$ is what links the said representations; where $^iX$ are the sprung mass coordinates.

The suspension force acting on each of the sprung masses is ${^iM} * {^i{\ddot{Z}}} + {^iB} * ({^i{\dot{Z}}}-{^i{\dot{z}}}) + {^iK} * ({^i{Z}}-{^i{z}})$; where, $^iZ$ and $^iz$ are the displacements of sprung and unsprung masses, and $^iB$ and $^iK$ are the damping and spring coefficients of $i$-th suspension, respectively.

The wheels of the vehicle are also modelled as rigid-bodies of mass $m$, acted upon by gravitational and suspension forces: ${^im} * {^i{\ddot{z}}} + {^iB} * ({^i{\dot{z}}}-{^i{\dot{Z}}}) + {^iK} * ({^i{z}}-{^i{Z}})$.

The tire forces are computed based on the friction curve for each tire: $\left\{\begin{matrix} {^iF_{t_x}} = F(^iS_x) \\{^iF_{t_y}} = F(^iS_y) \\ \end{matrix}\right.$; where, $^iS_x$ and $^iS_y$  are the longitudinal and lateral slips of $i$-th tire, respectively. Here, the friction curve is approximated by a two-piece cubic spline $F(S) = \left\{\begin{matrix} f_0(S); \;\; S_0 \leq S < S_e \\ f_1(S); \;\; S_e \leq S < S_a \\ \end{matrix}\right.$; where, $f_k(S) = a_k*S^3+b_k*S^2+c_k*S+d_k$ is a cubic polynomial function. The first segment of the said spline starts at zero $(S_0,F_0)$ and reaches the extremum point $(S_e,F_e)$, while the other segment starts at the extremum point $(S_e, F_e)$ and saturates at the asymptote point $(S_a, F_a)$, as shown in the inset of Fig. \hyperref[fig4]{\ref*{fig4}-A}.

Now, the tire slip is itself affected by various factors including tire stiffness $^iC_\alpha$, steering angle $\delta$, wheel speeds $^i\omega$, suspension forces $^iF_s$, and rigid-body momentum $^iP$, all of which affect the longitudinal/lateral/both components of the linear velocity of the vehicle. Longitudinal slip $^iS_x$ of $i$-th tire is computed by comparing the longitudinal components of surface velocity of $i$-th wheel (i.e., longitudinal linear velocity of vehicle) $v_x$ with the angular velocity $^i\omega$ of $i$-th wheel: ${^iS_x} = \frac{{^ir}*{^i\omega}-v_x}{v_x}$. Lateral slip $^iS_y$ of $i$-th tire is dependent on the direction the tire is pointing in and the direction it is moving in, commonly called the slip angle $\alpha$. It is computed by comparing the longitudinal $v_x$ (a.k.a. forward velocity) and lateral $v_y$ (a.k.a. side-slip velocity) components of vehicle's linear velocity: ${^iS_y} = tan(\alpha) = \frac{v_y}{\left| v_x \right|}$.

\subsection{Sensor Simulation}
\label{Sub-Section: Sensor Simulation}

The simulated vehicle is provided with the same sensing modalities as its real-world counterpart. Particularly, the throttle ($\tau$) and steering ($\delta$) sensors are simulated through a simple feedback loop.

The incremental encoders are simulated by measuring the rotation of rear wheels (i.e., output shaft of driving actuators): $^iN_{ticks} = {^iPPR} * {^iGR} * {^iN_{rev}}$; where, $^iN_{ticks}$ represents ticks measured by $i$-th encoder, $^iPPR$ is the base resolution (pulses per revolution) of $i$-th encoder, $^iGR$ is the gear-ratio of $i$-th motor, and $^iN_{rev}$ represents the number of revolutions of output shaft of $i$-th motor.

The IPS and IMU are simulated based on temporally-coherent rigid-body transform updates of the vehicle $\{v\}$ w.r.t. the world $\{w\}$: ${^w\mathbf{T}_v} = \left[\begin{array}{c | c} \mathbf{R}_{3 \times 3} & \mathbf{t}_{3 \times 1} \\ \hline \mathbf{0}_{1 \times 3} & 1 \end{array}\right] \in SE(3)$. The IPS provides 3-DOF positional coordinates $\{x,y,z\}$ of the vehicle, whereas the IMU provides linear accelerations $\{a_x,a_y,a_z\}$, angular velocities $\{\omega_x,\omega_y,\omega_z\}$ and 3-DOF orientation of the vehicle as Euler angles $\{\phi_x,\theta_y,\psi_z\}$ or quaternion $\{q_0,q_1,q_2,q_3\}$.

The LIDAR is simulated using iterative ray-casting \texttt{raycast}\{$^w\mathbf{T}_l$, $\vec{\mathbf{R}}$, $r_{max}$\} $\forall \; \theta \in \left [ \theta_{min}:\theta_{res}:\theta_{max} \right ]$ at $\sim$7 Hz update rate; where, ${^w\mathbf{T}_l} = {^w\mathbf{T}_v} * {^v\mathbf{T}_l} \in SE(3)$ is the relative transform of LIDAR \{$l$\} w.r.t. vehicle \{$v$\} w.r.t. world \{$w$\}, $\vec{\mathbf{R}} = \left [r_{max}*sin(\theta) \;\; r_{min}*cos(\theta) \;\; 0 \right ]^T$ is the direction vector of each ray-cast $R$, $r_{min}=$ 0.15 m and $r_{max}=$ 12 m are respectively the minimum and maximum linear ranges of LIDAR, $\theta_{min}=0^\circ$ and $\theta_{max}=360^\circ$ are respectively the minimum and maximum angular ranges of LIDAR, and $\theta_{res}=1^\circ$ is the angular resolution of LIDAR. The laser scan ranges are recorded by checking the ray-cast hits and thresholding the minimum linear range of LIDAR: \texttt{ranges[i]}$=\begin{cases} \texttt{hit.dist} & \text{ if } \texttt{ray[i].hit} \text{ and } \texttt{hit.dist} \geq r_{min} \\ \infty & \text{ otherwise} \end{cases}$; where, \texttt{ray.hit} is a Boolean flag that checks if a ray-cast hits any colliders in the scene and \texttt{hit.dist}$=\sqrt{(x_{hit}-x_{ray})^2 + (y_{hit}-y_{ray})^2 + (z_{hit}-z_{ray})^2}$ is the Euclidean distance from source of the ray-cast $\{x_{ray}, y_{ray}, z_{ray}\}$ to the hit-point $\{x_{hit}, y_{hit}, z_{hit}\}$.

The simulated physical cameras are parameterized by their focal length ($f=$ 3.04 mm), sensor size ($\{s_x, s_y\} = $ \{3.68, 2.76\} mm), target resolution (default = 720p) as well as distance to near and far clipping planes ($N=$ 0.01 m and $F=$ 1000 m). The viewport rendering pipeline for simulated cameras works in 3 stages. First, the camera view matrix $\mathbf{V} \in SE(3)$ is computed by taking the relative homogeneous transform of the camera $\{c\}$ w.r.t. the world $\{w\}$: $\mathbf{V} = \begin{bmatrix} r_{00} & r_{01} & r_{02} & t_{0} \\ r_{10} & r_{11} & r_{12} & t_{1} \\ r_{20} & r_{21} & r_{22} & t_{2} \\ 0 & 0 & 0 & 1 \\ \end{bmatrix}$; where, $r_{ij}$ and $t_i$ denote rotational and translational components, respectively. Next, the camera projection matrix $\mathbf{P} \in \mathbb{R}^{4 \times 4}$ is computed by projecting the world coordinates to image space coordinates: $\mathbf{P} = \begin{bmatrix} \frac{2*N}{R-L} & 0 & \frac{R+L}{R-L} & 0 \\ 0 & \frac{2*N}{T-B} & \frac{T+B}{T-B} & 0 \\ 0 & 0 & -\frac{F+N}{F-N} & -\frac{2*F*N}{F-N} \\ 0 & 0 & -1 & 0 \\ \end{bmatrix}$; where, $N$ and $F$ respectively denote distances to near and far clipping planes of the camera, and $L$, $R$, $T$ and $B$ respectively denote the left, right, top and bottom offsets of the sensor. The camera parameters $\{f,s_x,s_y\}$ are related to the projection matrix terms through the following relations: $f = \frac{2*N}{R-L}$, $a = \frac{s_y}{s_x}$, $\frac{f}{a} = \frac{2*N}{T-B}$. The perspective projection from the simulated camera's viewport is given by $\mathbf{C} = \mathbf{P}*\mathbf{V}*\mathbf{W}$; where, $\mathbf{C} = \left [x_c\;\;y_c\;\;z_c\;\;w_c \right ]^T$ represents the image space coordinates and $\mathbf{W} = \left [x_w\;\;y_w\;\;z_w\;\;w_w \right ]^T$ represents the world coordinates. Finally, this camera projection is converted into normalized device coordinates (NDC) by performing perspective divide (i.e., dividing throughout by $w_c$), obtaining a viewport projection by scaling and shifting the result and then using the rasterization process of the graphics API (e.g., DirectX for Windows, Metal for macOS and Vulkan for Linux). Additionally, a post-processing step simulates lens and film effects of the camera such as lens distortion, depth of field, exposure, ambient occlusion, contact shadows, bloom, motion blur, film grain, chromatic aberration, etc.

\subsection{Actuator Simulation}
\label{Sub-Section: Actuator Simulation}

The vehicle is actuated using two driving actuators and a steering actuator, the response delays and saturation limits of which are matched with their real-world counterparts by tuning their torque profiles and actuation limits, respectively.

The driving actuators drive the rear wheels by applying a torque: ${^i\tau_{drive}} = {^iI_w}*{^i\dot{\omega}_w}$; where, ${^iI_w} = \frac{1}{2}*{^im_w}*{^i{r_w}^2}$ is the moment of inertia, $^i\dot{\omega}_w$ is the angular acceleration,  $^im_w$ is the mass and $^ir_w$ is the radius of $i$-th wheel. Additionally, the holding torque of the driving actuators is simulated by applying an idle motor torque equivalent to the braking torque: ${^i\tau_{idle}} = {^i\tau_{brake}}$.

The front wheels are steered using a steering actuator, which produces a torque proportional to the required angular acceleration: $\tau_{steer} = I_{steer}*\dot{\omega}_{steer}$. The individual turning angles, $\delta_l$ and $\delta_r$, for left and right wheels, respectively, are calculated based on the commanded steering angle $\delta$, using the Ackermann steering geometry defined by wheelbase $l$ and track width $w$, as follows: $\left\{\begin{matrix} \delta_l = \textup{tan}^{-1}\left(\frac{2*l*\textup{tan}(\delta)}{2*l+w*\textup{tan}(\delta)}\right) \\ \delta_r = \textup{tan}^{-1}\left(\frac{2*l*\textup{tan}(\delta)}{2*l-w*\textup{tan}(\delta)}\right) \end{matrix}\right.$

\subsection{Infrastructure Simulation}
\label{Section: Infrastructure Simulation}

Simulated environments can be set-up in one of the following ways:
\begin{itemize}
     \item \textit{AutoDRIVE IDK:} The modular and reconfigurable infrastructure development kit (IDK) can be used to create custom scenarios and maps by setting up the terrain modules, road networks, obstruction modules and traffic elements. These assets are present within the simulator source files. Particularly, preconfigured maps depicted in Fig. \hyperref[fig3]{\ref*{fig3}-F-(i)}, Fig. \hyperref[fig3]{\ref*{fig3}-F-(iii)} and Fig. \hyperref[fig3]{\ref*{fig3}-F-(iv)} have been constructed using the AutoDRIVE IDK.
     \item \textit{Plug-In Scenarios:} AutoDRIVE Simulator supports third-party tools (e.g., RoadRunner \cite{RoadRunner2021}) and modular open-source architecture (MOSA) standards (e.g., OpenSCENARIO \cite{OpenSCENARIO2021}, OpenDRIVE \cite{OpenDRIVE2021}, etc.) that enable extensibility. Additionally, users can import a wide array of plugins, packages and assets in a variety of industry-standard formats (e.g., FBX, OBJ, SKP, 3DS, USD, etc.) for developing or customizing driving scenarios. Furthermore, graphics textures designed for AutoDRIVE Testbed can be first imported into the simulator before large-scale printing and setup in real-world. Particularly, preconfigured map depicted in Fig. \hyperref[fig3]{\ref*{fig3}-F-(ii)} has been designed using a third-party graphics editing software, imported in AutoDRIVE Simulator and finally printed and setup using AutoDRIVE Testbed.
     \item \textit{Unity Terrain:} Being built atop the Unity game engine, AutoDRIVE Simulator natively supports scenario design and development using Unity Terrain \cite{UnityTerrain2021}. Users can define the terrain mesh, texture, heightmap, vegetation, skybox, wind, etc. to design on-road/off-road scenarios and perform variability testing.
\end{itemize}

At every time step, the simulator performs mesh-mesh interference detection and computes the contact forces, frictional forces and momentum transfer, along with linear and angular drag acting on the rigid-bodies (refer Fig. \hyperref[fig4]{\ref*{fig4}-B}).

\subsection{Simulator Features}
\label{Section: Simulator Features}

\begin{table*}[t]
\centering
\caption{Decision matrix for choosing the demonstration case-studies}
\label{tab2}
\resizebox{\textwidth}{!}{%
\begin{tabular}{llllllll}
\rowcolor[HTML]{85DFFF} 
\multicolumn{1}{c}{\cellcolor[HTML]{85DFFF}\textbf{Autonomy Algorithm}} & \multicolumn{1}{c}{\cellcolor[HTML]{85DFFF}\textbf{Platform Exploited}}          & \multicolumn{1}{c}{\cellcolor[HTML]{85DFFF}\textbf{Development Framework}}    & \multicolumn{1}{c}{\cellcolor[HTML]{85DFFF}\textbf{Autonomy Stack}}                    & \multicolumn{1}{c}{\cellcolor[HTML]{85DFFF}\textbf{Science and Technology Demonstrated}}                                                                                                     & \multicolumn{1}{c}{\cellcolor[HTML]{85DFFF}\textbf{Agents Involved}} & \multicolumn{1}{c}{\cellcolor[HTML]{85DFFF}\textbf{Sensors Employed}}      & \multicolumn{1}{c}{\cellcolor[HTML]{85DFFF}\textbf{Actuators Controlled}}       \\
\cellcolor[HTML]{C0C0C0}Autonomous Parking                              & AutoDRIVE Testbed                                                                & \begin{tabular}[c]{@{}l@{}}AutoDRIVE ROS Package\\ (Python, C++)\end{tabular} & \begin{tabular}[c]{@{}l@{}}Modular (Perception,\\ Planning and Control)\end{tabular}   & \begin{tabular}[c]{@{}l@{}}Teleoperation, SLAM, Probabilistic\\ Map-Based Localization, Global Planning,\\ Local Planning, Motion Control,\\ Static/Dynamic Collision Avoidance\end{tabular} & Single-Agent System                                                  & LIDAR                                                                      & \begin{tabular}[c]{@{}l@{}}Driving Actuators,\\ Steering Actuator\end{tabular}  \\
\cellcolor[HTML]{C0C0C0}Behavioral Cloning                              & \begin{tabular}[c]{@{}l@{}}AutoDRIVE Simulator,\\ AutoDRIVE Testbed\end{tabular} & \begin{tabular}[c]{@{}l@{}}AutoDRIVE Python API\\ (Python)\end{tabular}       & \begin{tabular}[c]{@{}l@{}}End-to-End (Sensorimotor\\ Policy)\end{tabular}             & \begin{tabular}[c]{@{}l@{}}Computer Vision, Deep Imitation Learning,\\ Lane Keeping, Sim2Real Transition\end{tabular}                                                                        & Single-Agent System                                                  & Front Camera                                                               & \begin{tabular}[c]{@{}l@{}}Driving Actuators,\\ Steering Actuator\end{tabular}  \\
\cellcolor[HTML]{C0C0C0}Intersection Traversal                          & AutoDRIVE Simulator                                                              & ML-Agents Toolkit (C\#)                                                         & \begin{tabular}[c]{@{}l@{}}End-to-End (Sensorimotor\\ Policy)\end{tabular}             & \begin{tabular}[c]{@{}l@{}}V2V Communication, Deep Reinforcement\\ Learning, Dynamic Collision Avoidance,\\ Multi-Agent Cooperation and Coordination\end{tabular}                            & Multi-Agent System                                                   & \begin{tabular}[c]{@{}l@{}}Incremental\\ Encoders,\\ IPS, IMU\end{tabular} & \begin{tabular}[c]{@{}l@{}}Steering Actuator\\ (Constant Throttle)\end{tabular} \\
\cellcolor[HTML]{C0C0C0}Smart City Management                           & AutoDRIVE Simulator                                                              & \begin{tabular}[c]{@{}l@{}}AutoDRIVE Webapp API\\ (Python)\end{tabular}       & \begin{tabular}[c]{@{}l@{}}Modular (Surveillance,\\ Planning and Control)\end{tabular} & \begin{tabular}[c]{@{}l@{}}V2I Communication, IoT, Centralized Control\\ and Coordination\end{tabular}                                                                                       & Single-Agent System                                                  & None                                                                       & \begin{tabular}[c]{@{}l@{}}Driving Actuators,\\ Steering Actuator\end{tabular} 
\end{tabular}
}
\end{table*}

AutoDRIVE Simulator is developed atop Unity \cite{Unity2021} game engine, which employs PhysX \cite{PhysX2021} for simulating multi-threaded framerate-independent kinematics and dynamics of all the physical entities, and exploits High Definition Render Pipeline (HDRP) \cite{HDRP2021} along with Post-Processing Stack \cite{PPS2021} for rendering photorealistic graphics.

The simulator features an interactive graphical user interface (GUI) consisting of Menu Panel on left-hand side and Heads-Up Display (HUD) on right-hand side. Fig. \hyperref[fig4]{\ref*{fig4}-C} depicts the simulator's GUI with both the panels enabled. The Menu Panel hosts input fields and buttons to configure and control various features of the simulator (refer Fig. \hyperref[fig4]{\ref*{fig4}-D}). This includes controls for the communication bridge, along with a series of buttons for (a) toggling between manual and autonomous driving modes for the ego vehicle; (b) switching between the available scene cameras, each providing a distinct view; (c) altering the graphics quality to match the quality-performance trade-off; (d) toggling the scene light to simulate day and night driving conditions; (e) resetting the scene to initial conditions; and (f) quitting the simulator application. The HUD Panel, on the other hand, displays prominent simulation parameters along with vehicle status and sensory data in real-time. It also hosts a time-synchronized data recording functionality, which can be used to export vehicle as well as infrastructure data for a specific run, thereby fostering data-driven approaches aimed at autonomous driving and smart city management.

The simulator natively supports C\# scripting, which can be leveraged to customize existing and/or introduce new features, functionalities, modules, behaviors, physics, graphics, communication bridges, APIs and even setup co-simulation frameworks with other simulation tools.

Finally, it is worth mentioning that the simulator, in its source form, has been integrated with various plugins and packages such as the Unity ML Agents Toolkit \cite{MLAgents2018}, a machine learning framework for developing and deploying deep imitation/reinforcement learning-based applications directly from within the simulator.

\section{AutoDRIVE Devkit}
\label{Section: AutoDRIVE Devkit}

AutoDRIVE Devkit is a collection of software packages, application programming interfaces (APIs) and tools, which enables flexible development of autonomous driving as well as smart city management algorithms targeted towards the testbed and/or simulator. It supports both local as well as distributed computing, thereby allowing development of both centralized and decentralized autonomy algorithms.

\subsection{Autonomous Driving Software Stack}
\label{Sub-Section: ADSS}

The autonomous driving software stack (ADSS) aids in development of autonomy algorithms specifically targeting the vehicle. It can be used to develop single as well as multi-agent autonomous driving algorithms.

\subsubsection{ROS Package}
\label{Sub-Sub-Section: ROS Package}

AutoDRIVE ROS package supports flexible development of modular autonomy algorithms. It can be installed on a ROS-compatible workstation for interfacing with the simulator (locally/remotely), or directly on Nigel's on-board computer for hardware deployment.

\subsubsection{Scripting APIs}
\label{Sub-Sub-Section: Scripting APIs}

AutoDRIVE Devkit currently offers scripting APIs for Python and C++, which can be exploited to develop high-performance autonomy algorithms, without ROS as an intermediary. Such source codes can be interfaced with the simulator (locally/remotely), or directly deployed on Nigel's on-board computer for hardware validation.

\subsection{Smart City Software Stack}
\label{Sub-Section: SCSS}

The smart city software stack (SCSS) aids in development of autonomy algorithms specifically targeting the infrastructure. It can work in tandem with ADSS to develop smart city applications pertaining to traffic management.

\subsubsection{SCM Server}
\label{Sub-Sub-Section: SCM Server}

AutoDRIVE Devkit offers a centralized Smart City Manager (SCM) server to monitor and control various ``smart'' elements. The server hosts a database to keep track of all the vehicles along with active and passive traffic elements within a particular scene.

\subsubsection{SCM Webapp}
\label{Sub-Sub-Section: SCM Webapp}

AutoDRIVE SCM hosts an interactive webapp, which allows the users to connect with the database for monitoring and controlling the traffic flow in real-time.


\section{Demonstration Case-Studies}
\label{Section: Demonstration Case-Studies}

\begin{figure*}[t]
\centering
\includegraphics[width=\linewidth]{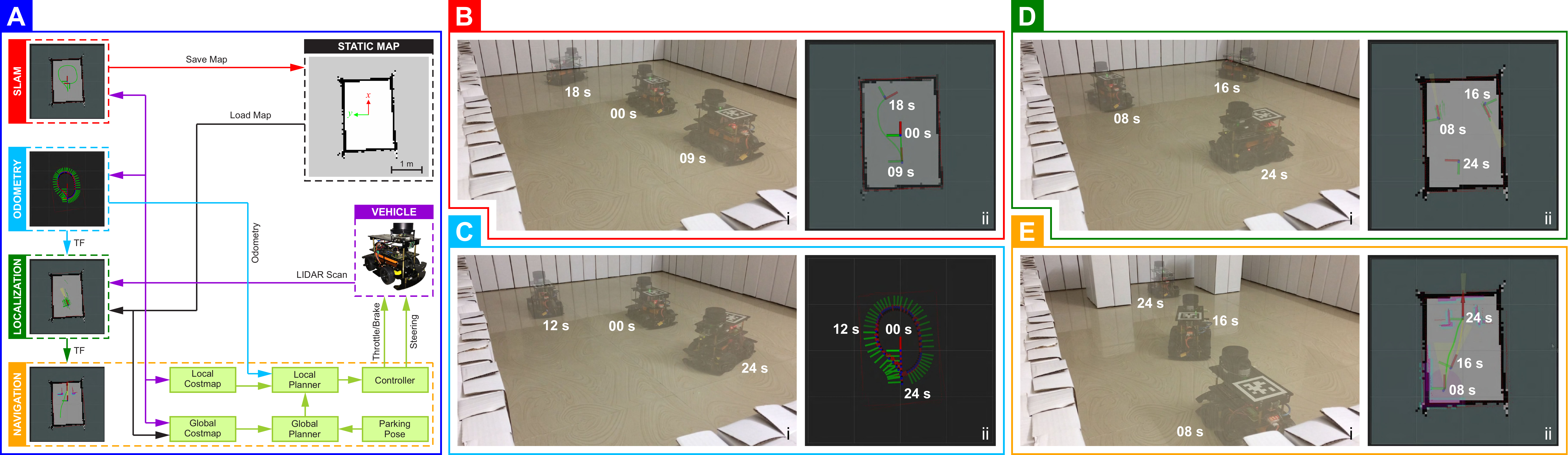}
\caption{Autonomous parking: [A] high-level architecture of the autonomy algorithm; [B], [C] [D] and [E] respectively depict temporal analysis of simultaneous localization and mapping, odometry, localization, and navigation modules - (i) physical vehicle driving in real-world setting, (ii) visualization of software algorithm; notice the additional boxes acting as unmapped obstacles in [E]. Video: \url{https://youtu.be/oBqIZZA0wkc}}
\label{fig5}
\end{figure*}

\begin{figure*}[t]
\centering
\includegraphics[width=\linewidth]{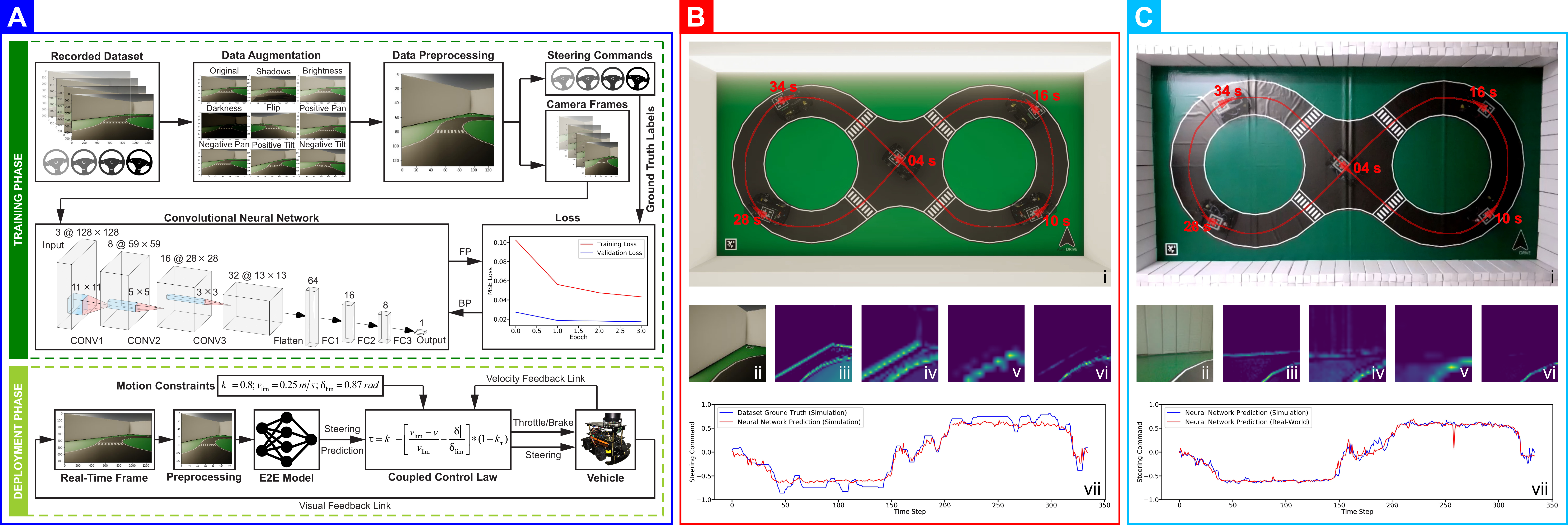}
\caption{Behavioral cloning: [A] high-level architecture of the training and deployment pipelines; [B] and [C] respectively depict behavioral analysis of the autonomous vehicle in virtual and real world settings - (i) trajectory tracked by the vehicle, (ii) a sample pre-processed camera frame fed as input to the neural network, (iii), (iv) and (v) respectively depict activation maps of first, second and third convolutional layers of the neural network, (vi) salient activations from all the activation maps, (vii) neural network prediction analysis for one complete lap. Video: \url{https://youtu.be/rejpoogaXOE}}
\label{fig6}
\end{figure*}

This work showcases key features and capabilities of the AutoDRIVE Ecosystem through 4 carefully shortlisted case-studies (refer Table \ref{tab2}). Although this paper cannot furnish exhaustive details pertaining to any particular demonstration, we recommend interested readers to peruse this technical report \cite{AutoDRIVEReport2021}.

It is to be noted that the presented demonstrations are by no means exhaustive and that AutoDRIVE Ecosystem can be employed to develop, simulate and deploy a much wider array of applications including (but not limited to) synthetic/real/hybrid data collection and labeling; traditional (e.g., deterministic/probabilistic, classical/optimal, etc.) as well as modern (e.g., deep imitation/reinforcement/hybrid learning, etc.) algorithms for perception, state estimation, path/motion planning and motion control; modular as well as end-to-end autonomy stacks; benchmarking existing solutions (education) or innovating novel scientific and technological approaches for autonomy (research), etc.

\subsection{Autonomous Parking}
\label{Sub-Section: Autonomous Parking}

This demonstration leveraged AutoDRIVE's ROS-enabled capabilities to demonstrate autonomous parking (refer Fig. \hyperref[fig5]{\ref*{fig5}-A}). First, the vehicle mapped its surroundings using Hector SLAM algorithm \cite{HectorSLAM2011} (refer Fig. \hyperref[fig5]{\ref*{fig5}-B}). It could then localize itself against this known static map using range-flow-based odometry \cite{RF2O2016} (refer Fig. \hyperref[fig5]{\ref*{fig5}-C}) and adaptive particle filter algorithm \cite{AMCL2001} (refer Fig. \hyperref[fig5]{\ref*{fig5}-D}). For autonomous navigation, the vehicle planned a feasible global path from its current pose to parking pose using A* algorithm \cite{AStar1968}, while also re-planning its local trajectory for dynamic collision avoidance using timed-elastic-band approach \cite{TEBPlanner2017}. A proportional controller generated driving (throttle/brake) and steering commands for the vehicle to track the local trajectory (refer Fig. \hyperref[fig5]{\ref*{fig5}-E}). Future work in this direction can include benchmarking of various SOTA and/or novel algorithms for mapping, localization, path planning and motion control.

\begin{figure*}[t]
\centering
\includegraphics[width=\linewidth]{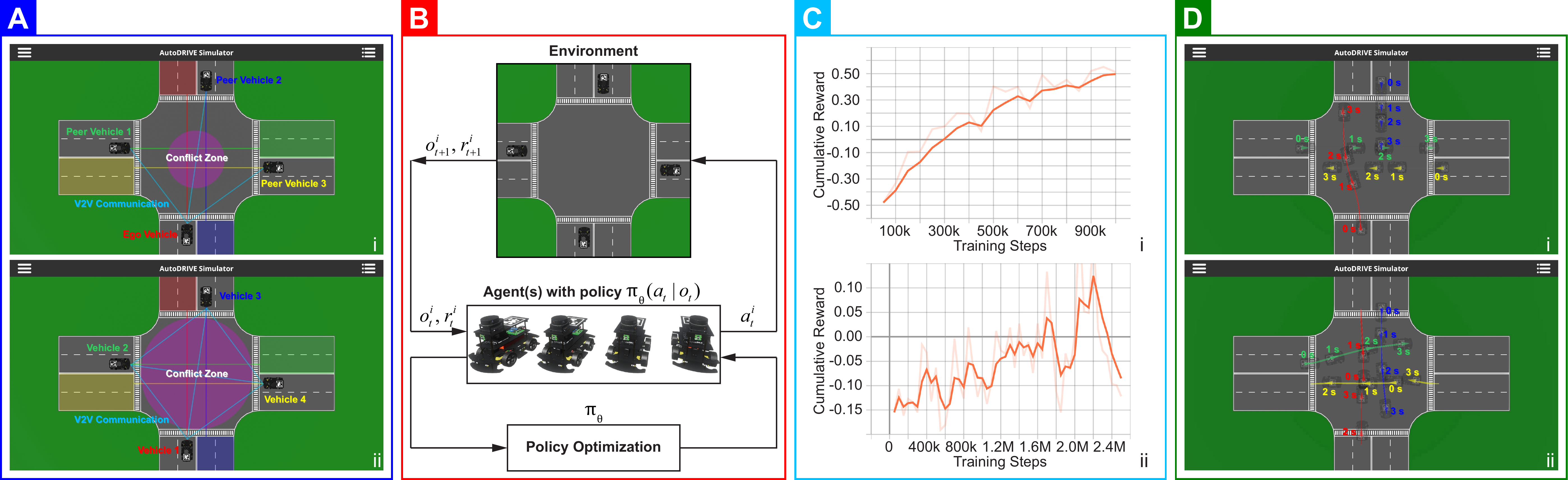}
\caption{Intersection traversal: [A] learning scenario descriptions; [B] deep reinforcement learning architecture; [C] and [D] respectively depict training and deployments results - (i) single-agent learning scenario, (ii) multi-agent learning scenario. Video: \url{https://youtu.be/AEFJbDzOpcM}}
\label{fig7}
\end{figure*}

\begin{figure*}[t]
\centering
\includegraphics[width=\linewidth]{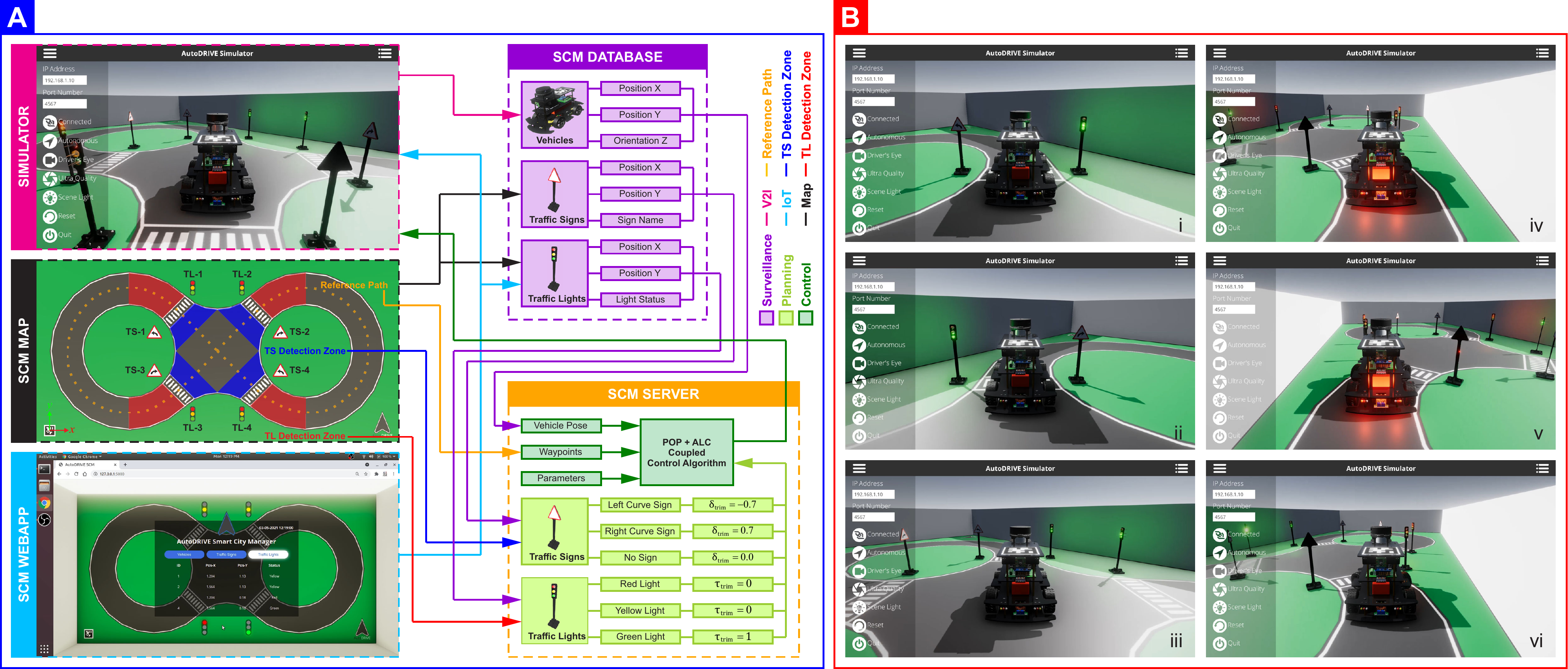}
\caption{Smart city management: [A] high-level architecture of the autonomy algorithm; [B] snapshot instances from simulation - (i) vehicle observing left-curve sign, (ii) vehicle observing right-curve sign, (iii) vehicle crossing the intersection, (iv) vehicle stopping at red light, (v) vehicle stopping at yellow light, (vi) vehicle resuming on green light; the traffic lights are toggled manually. Video: \url{https://youtu.be/fnxOpV1gFXo}}
\label{fig8}
\end{figure*}

\subsection{Behavioral Cloning}
\label{Sub-Section: Behavioral Cloning}

This demonstration was based on \cite{RBC2021}, wherein the objective was to employ a convolutional neural network (CNN) \cite{Krizhevsky2012} for cloning the end-to-end driving behavior of a human. As such, AutoDRIVE Simulator was exploited to record 5-laps worth of temporally-coherent labeled manual driving data, which was balanced, augmented and pre-processed using standard computer vision techniques, to train a 6-layer deep CNN (refer Fig. \hyperref[fig6]{\ref*{fig6}-A}). Post training for 4 epochs, with a learning rate of 1e-3 using the Adam optimizer \cite{Kingma2014}, the training and validation losses converged stably without over/under-fitting. The trained model was then deployed back into the virtual world to validate its performance through activation and prediction analyses (refer Fig. \hyperref[fig6]{\ref*{fig6}-B}). Further, the same model was transferred to AutoDRIVE Testbed for validating sim2real capability of the ecosystem through activation and prediction analyses (refer Fig. \hyperref[fig6]{\ref*{fig6}-C}). In order to have a zero-shot sim2real transition: (i) the physical and visual aspects of virtual world were setup as close to the real world as possible, and (ii) the exhaustive data augmentation pipeline implicitly performed domain randomization. However, further investigation is required to comment on and improve the robustness of sim2real transfer considering sensor simulation, vehicle modeling and scenario representation. Finally, although this work adopted a coupled-control law for vehicle motion smoothing, a potential improvement could be to investigate independent actuation smoothing techniques like low-pass filters or proximal bounds.

\subsection{Intersection Traversal}
\label{Sub-Section: Intersection Traversal}

Inspired from \cite{MARL2020}, this work demonstrates single and multi-agent (refer Fig. \hyperref[fig7]{\ref*{fig7}-A}) intersection traversal using deep reinforcement learning (refer Fig. \hyperref[fig7]{\ref*{fig7}-B}). Each agent collected a vectorized observation $o_{t}^{i} = \left [ g^{i}, \tilde{p}^{i}, \tilde{\psi}^{i}, \tilde{v}^{i} \right ]_{t}$, including its relative goal location $g_{t}^{i}$, along with relative location $\tilde{p}_{t}^{i}$, relative yaw $\tilde{\psi}_{t}^{i}$ and velocity $\tilde{v}_{t}^{i}$ of its peers obtained through V2V communication. The action space of each agent, $a_{t}^{i}$, included discretized steering $\delta_{t}^{i} \in \left \{ -1, 0, 1 \right \}$ and constant throttle ($\tau_{t}^{i}=80\%$). An extrinsic reward function $r_{t}^{i} = \begin{cases} r_{goal} = +1\\ r_{collision} = -0.425 * \left \| g_{t}^{i} \right \|_{2} \end{cases}$ kept agents in check while training a 3-layer fully-connected neural network based policy, $\pi_\theta \left ( a_t | o_t \right )$, using PPO algorithm \cite{PPO2017} (refer Fig. \hyperref[fig7]{\ref*{fig7}-C}). This ultimately resulted in all agents being able to traverse the intersection safely (refer Fig. \hyperref[fig7]{\ref*{fig7}-D}). Although we could not implement this application in real-world owing to monetary constraints, investigating the sim2real transition of this application would be a natural progression of this work. Additionally, application of actuation smoothing techniques could be a potential improvement.

\subsection{Smart City Management}
\label{Sub-Section: Smart City Management}

This novel use case of smart city traffic management was possible due to AutoDRIVE's V2I and IoT capabilities. As depicted in Fig. \hyperref[fig8]{\ref*{fig8}-A}), the SCM server hosted a database to keep track of all the traffic elements, and acted as a high-level behavior planner for the ego vehicle. It switched the vehicle behavior upon detecting respective traffic signs and lights by setting the appropriate throttle and steering trims, which were then passed on to proximally optimal predictive (POP) controller coupled with adaptive longitudinal controller (ALC) \cite{POP2022}. Finally, SCM server teleoperated the vehicle to achieve the mission objective (refer Fig. \hyperref[fig8]{\ref*{fig8}-B}). A natural progression of this work would be to implement a multi-agent scenario, preferably within a mixed-reality digital-twin setting, to investigate different strategies for efficient smart city traffic management.


\section{Summary}
\label{Section: Summary}

AutoDRIVE was developed with an aim of tightly integrating real and virtual worlds into a common toolchain, without compromising on the comprehensiveness, flexibility and accessibility required for prototyping and validating autonomy solutions. It has numerous applications, which are bound to increase as the ecosystem is upgraded. Potential improvements include supporting heterogeneous vehicles and robotic pedestrians, full-scale vehicles and environments, expanding API support and adding extended reality capabilities, to name a few. We hope that the community benefits from adopting this ecosystem, may it be for education, research or anything in between.


\section{Supplemental Material}
\label{Section: Supplemental Material}

AutoDRIVE Ecosystem is openly accessible.
\begin{itemize}
    \item \textbf{AutoDRIVE Ecosystem Website:}\\{\footnotesize \url{https://AutoDRIVE-Ecosystem.github.io}}
    \item \textbf{AutoDRIVE Ecosystem GitHub Organization:}\\{\footnotesize \url{https://github.com/AutoDRIVE-Ecosystem}}
    \item \textbf{AutoDRIVE Ecosystem YouTube Channel:}\\{\footnotesize \url{https://www.youtube.com/@AutoDRIVE-Ecosystem}}
\end{itemize}



\balance
\bibliographystyle{IEEEtran}
\bibliography{References}

\end{document}